\documentclass[10pt,twocolumn,final]{IEEEtran}
\usepackage{times}
\usepackage{graphics}
\usepackage{graphicx}
\usepackage{amsfonts}
\usepackage{amsmath}
\usepackage{amssymb}
\usepackage{booktabs}
\usepackage{mathrsfs}
\usepackage{amsthm}
\usepackage{color}
\usepackage{algorithm}
\usepackage{algpseudocode}
\usepackage{subcaption}
\usepackage{cite}
\usepackage{tabularx}
\usepackage{mathtools}
\usepackage{bbm}
\usepackage{multirow}
\usepackage[numbers,sort&compress]{natbib}
\usepackage{url}

\begin{document}

\title{Probabilistic Deep Metric Learning for Hyperspectral Image Classification}

\author{Chengkun Wang, Wenzhao Zheng, \\
Xian~Sun,~\IEEEmembership{Senior Member,~IEEE},  Jiwen~Lu,~\IEEEmembership{Senior Member,~IEEE}, and Jie~Zhou,~\IEEEmembership{Senior Member,~IEEE}\IEEEcompsocitemizethanks{\IEEEcompsocthanksitem

Chengkun Wang, Wenzhao Zheng, Jiwen Lu, and Jie Zhou are with the Department of Automation, Tsinghua University, China. State Key Lab of Intelligent Technologies and Systems, Tsinghua University, China. Beijing National Research Center for Information Science and Technology, China. Email: wck20@mails.tsinghua.edu.cn; zhengwz18@mails.tsinghua.edu.cn; lujiwen@tsinghua.edu.cn;  jzhou@tsinghua.edu.cn.

Xian Sun is with Aerospace Information Research Institute, Chinese Academy of Sciences, Beijing 100190, China, also with the School of Electronic, Electrical and Communication Engineering, University of Chinese Academy of Sciences, Beijing 100190, China, and also with the Key Laboratory of Network Information System Technology (NIST), Aerospace Information Research Institute, Chinese Academy of Sciences, Beijing 100190, China. Email: sunxian@mail.ie.ac.cn.}}

\IEEEcompsoctitleabstractindextext{

}

\maketitle

\IEEEpeerreviewmaketitle

\begin{abstract}
This paper proposes a probabilistic deep metric learning (PDML) framework for hyperspectral image classification, which aims to predict the category of each pixel for an image captured by hyperspectral sensors.
The core problem for hyperspectral image classification is the spectral variability between intraclass materials and the spectral similarity between interclass materials, motivating the further incorporation of spatial information to differentiate a pixel based on its surrounding patch.
However, different pixels and even the same pixel in one patch might not encode the same material due to the low spatial resolution of most hyperspectral sensors, leading to an inconsistent judgment of a specific pixel. 
To address this issue, we propose a probabilistic deep metric learning framework to model the categorical uncertainty of the spectral distribution of an observed pixel. 
We propose to learn a global probabilistic distribution for each pixel in the patch and a probabilistic metric to model the distance between distributions.
We treat each pixel in a patch as a training sample, enabling us to exploit more information from the patch compared with conventional methods.
Our framework can be readily applied to existing hyperspectral image classification methods with various network architectures and loss functions. 
Extensive experiments on four widely used datasets including IN, UP, KSC, and Houston 2013 datasets demonstrate that our framework improves the performance of existing methods and further achieves the state of the art. 
Code is available at: \url{https://github.com/wzzheng/PDML}.
\end{abstract}
\IEEEdisplaynotcompsoctitleabstractindextext
\section{introduction}
Hyperspectral remote sensing (HSRS) has attracted wide attention in recent years, which is able to capture abundant information about the landform with hundreds of fine-grained and continuous spectral bands~\cite{landgrebe2002hyperspectral,ghamisi2017advances,bioucas2013hyperspectral}. 
As one of the most important tasks in HSRS, hyperspectral image classification (HSIC) aims to assign each pixel with one label from a pre-defined set, such as gravel and water~\cite{he2017recent,zhu2020residual}. 
HSIC has been widely employed in various situations including crop discrimination~\cite{eddy2014weed}, military reconnaissance~\cite{khan2018modern}, and geological examination~\cite{villa2011hyperspectral}.

\begin{figure}[t]
\centering
\includegraphics[width=0.475\textwidth]{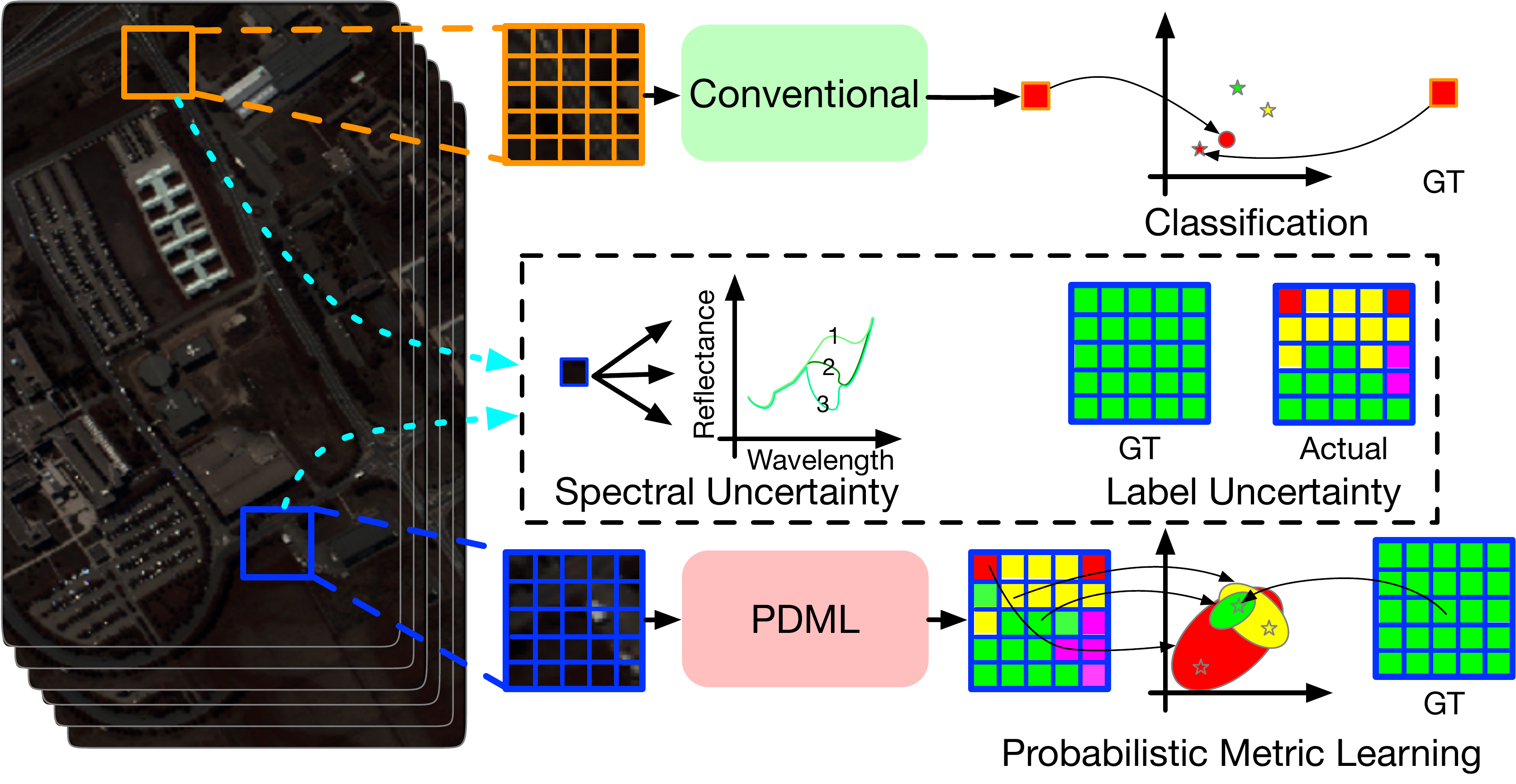}
\caption{The motivation of the proposed PDML framework. 
Existing HSIC methods take as input an image patch centered at the concerned pixel and output a single classification score for the center pixel, which suffers from the inaccuracy of the training signals caused by the spectral uncertainty (i.e., intraclass variability and interclass similarity) and the label uncertainty (i.e., mixed pixels and lack of labels for surrounding pixels). 
The proposed PDML framework employs an encoder-decoder to model the uncertainty of each pixel by a probabilistic distribution with different variances. 
We further adopt a probabilistic deep metric learning method to constrain the distances between distributions, which can suppress the influence of the uncertainty in training samples to obtain a model more tolerant to the possible inconsistency in the training data.
(Best viewed in color.)
}
\label{fig:motivation}
\vspace{-5mm}
\end{figure}

A core issue for hyperspectral image classification is the spectral uncertainty including the spectral variability between intraclass materials and the spectral similarity between interclass materials~\cite{zare2013endmember, chang2000information}. 
The spectral uncertainty results from a number of factors, such as natural spectrum variation, instrument noises, atmospheric effects, etc.~\cite{he2017recent, shaw2002signal}. 
Under such circumstances, the spectral characteristics alone are incapable of accurately and robustly discriminating a certain pixel in a hyperspectral image.
This motivates the recent spectral-spatial classification methods to additionally take spatial information into consideration in order to reduce the effect of spectral uncertainty~\cite{sun2014supervised,paoletti2019deep,fang2014spectral,kang2013spectral}. 
They employ spatial neighborhoods of the concerned pixel as well as their spectral features and take the resulting 3D patch as the input with a Convolutional Neural Network (CNN) to process hyperspectral images~\cite{hamida20183,chen2016deep,article,li2017spectral}. 
Essentially, these methods perform multiple sampling of adjacent pixels and employ a voting scheme to enhance the confidence of the overall prediction, where the voting weights are implicitly learned in the CNNs.
The improvement is effective assuming that adjacent pixels capture the same type of materials. 
However, this assumption might not hold due to the native low spatial resolution of hyperspectral sensors~\cite{audebert2019deep,zhang2018diverse}, making surrounding pixels possible to encoder features from different materials.
Furthermore, even a single pixel might contain different materials (i.e., mixed pixel)~\cite{rajabi2015sparsity,hsieh2001effect,xie2018unsupervised}.
This brings uncertainty to the labels of both surrounding pixels and center pixels (i.e., label uncertainty), leading to the inaccuracy of the supervisory signal.

In this paper, we propose a probabilistic deep metric learning (PDML) framework to address the aforementioned spectral uncertainty and label uncertainty, as demonstrated in Figure~\ref{fig:motivation}. 
We employ a high-dimension probabilistic distribution to represent a pixel to directly model the uncertainty. 
Different from existing methods which learn a single deterministic representation only for the concerned center pixel, we use an encoder-decoder architecture to learn a probabilistic representation for each pixel in the patch.
We similarly assign the same label to all the pixels due to the lack of ground truth labels for the surrounding pixels but accompany them with different degrees of uncertainty.
We achieve this by enforcing a larger variance for a further pixel since it is more probable to capture a different material from the center pixel. 
We further propose a deep probabilistic metric learning method to constrain the similarities between probabilistic distributions to increase interclass variance and decrease intraclass variance.
We employ Monte Carlo sampling to generate a number of samples following the corresponding distribution to represent each pixel and apply a metric learning loss to them. 
The overall framework can be learned simultaneously in an end-to-end manner during training, and we only extract the mean vector of the center pixel for classification during inference, introducing no additional workload compared with existing methods.
The proposed PDML framework directly models the implicit uncertainty within the hyperspectral image patches, enabling our model to adaptively rectify the influence of possible inconsistent training samples.
Furthermore, we treat all the pixels in a patch as training samples with the same label but different variances and impose constraints on all of them to exploit full information from the patch, compensating for the lack of training data which is a common problem in HSIC.
Extensive experiments on four widely used datasets demonstrate the effectiveness of the proposed PDML framework.

In general, we summarize our key contributions as follows:

1) To the best of our knowledge, we are the first to consider both the spectral uncertainty and label uncertainty in a hyperspectral image and propose to use a probabilistic embedding to represent each pixel.

2) We propose a probabilistic deep metric learning method, where we generate various samples following the distribution of each pixel by Monte Carlo sampling and impose a metric learning loss to enlarge the interclass distances as well as reduce the intraclass distances in the embedding space.

3) We conduct experiments on four widely used hyperspectral datasets, which demonstrates that our proposed framework can be applied to various existing methods to boost their performance and further achieve the state-of-the-art.
\section{Related Works}
In this section, we review recent advancements in three related areas: deep metric learning, probabilistic embedding learning, and hyperspectral image classification.
\subsection{Deep Metric Learning}
Metric learning aims to construct a discriminative distance function to effectively represent the similarities between samples. 
Deep metric learning employs deep networks to transform images to an embedding space with the objective to decrease intraclass distances and increase interclass distances~\cite{wang2022introspective,zheng2022dynamic,zheng2021deep1,zheng2021deep2,zhang2022attributable,zheng2020structural}. 
Numerous deep metric learning methods focus on designing discriminative losses to instantiate this objective~\cite{cakir2019deep,sohn2016improved,song2016deep,wang2019multi,yu2019deep}. 
For example, the contrastive loss~\cite{hadsell2006dimensionality} pulls samples from the same class as close as possible and pushes away samples of different categories by at least a fixed margin. 
The triplet loss is more flexible which only enforces a distance ranking within triplets~\cite{schroff2015facenet}. 
Furthermore, Wang~\emph{et al.}~\cite{wang2019multi} proposed a multi-similarity loss to consider three types of similarities between samples.

In addition to the loss functions, how to effectively construct training tuples significantly affects the performance of deep metric learning~\cite{wu2017sampling,duan2018deep,zheng2019hardness,harwood2017smart,xuan2020improved,zheng2020deep}. 
For example, Harwood~\emph{et al.}~\cite{harwood2017smart} improved the widely used hard negative mining strategy~\cite{schroff2015facenet} and proposed a smart mining strategy to choose challenging negative samples adaptively considering the current training stage.
 In addition, Wu~\emph{et al.}~\cite{wu2017sampling} uniformly selected samples based on the distance distributions and Xuan~\emph{et al.}~\cite{xuan2020improved} claimed that sampling easy positive samples is helpful to the generalization performance.

\subsection{Probabilistic Embedding Learning}
Probabilistic embedding learning was initially proposed and developed for word embedding since it can naturally model the inherent hierarchies and ambiguity in language.
It is effectively applied to various natural language processing tasks~\cite{vilnis2014word,li2018smoothing,li2020adaptive,nguyen2017mixture}. 
For example, Vilnis~\emph{et al.}~\cite{vilnis2014word} introduced the Gaussian embedding for word representation. 
Li~\emph{et al.}~\cite{li2020adaptive} proposed a probabilistic embedding learning approach that adaptively adjusts and updates word embeddings. 
Probabilistic embedding learning is also proven to be beneficial to computer vision tasks, such as face recognition~\cite{shi2019probabilistic,chang2020data}, pose estimation~\cite{sun2020view}, cross-modal retrieval~\cite{chun2021probabilistic}, and regression problems~\cite{li2021learning}. 
Recently, probabilistic embedding has been introduced to metric learning by Oh~\emph{et al.}~\cite{oh2018modeling} who proposed a hedged instance embedding learning method to address the semantic ambiguity of natural images.

Our work further extends probabilistic embedding learning to hyperspectral image classification.
We model the spectral and label uncertainty in image patches using Gaussian distributions and propose a probabilistic metric learning framework to constrain the distributions. 
Different from existing works, we impose an uncertainty prior to the variance of the distributions so that the variance of outer pixels in a patch is larger than that of inner pixels. 
To the best of our knowledge, PDML is the first work that employs probabilistic embedding to model the spectral and label uncertainty in hyperspectral remote sensing.

\subsection{Hyperspectral Image Classification}
Hyperspectral Image Classification (HSIC) aims to assign the correct semantic label to each pixel in the hyperspectral images. 
Benifiting from the great power of deep neural networks, deep-learning-based methods~\cite{9785505,9782104,9585383,9693311,8463629,zhan2017semisupervised,zhu2020residual,sun2019spectral,yang2018hyperspectral,yu2021feedback,he2018feature} significantly outperform conventional methods~\cite{blanzieri2008nearest,sun2014active,mianji2011robust,li2011spectral,li2013generalized} and have dominated the HSIC fields. 
For example, pioneer methods employed 1-D CNN~\cite{hu2015deep}, recurrent neural networks (RNN)~\cite{mou2017deep}, and generative adversarial networks~\cite{zhan2017semisupervised} to capture abundant information from spectral bands. 
Subsequent methods began to exploit the spatial information to address the spectral variability of hyperspectral images. 
For example, Yang~\emph{et al.}~\cite{yang2018hyperspectral} proposed a 3D structure to simultaneously consider the spectral-spatial information using image patches as inputs.
He~\emph{et al.}~\cite{he2018feature} proposed to use the multiscale covariance maps for better extraction of spatial information. 
Recently, a number of methods employed the attention-based networks for the HSIC task and achieved excellent performance~\cite{9693311,yu2021feedback,sun2019spectral}.

In addition, metric learning has been applied to hyperspectral classification~\cite{deng2019deep,cao2019hyperspectral,dong2017dimensionality} for better discriminative representations of pixels.
Compared with conventional deep-learning-based methods, they further constrained the relations between patches and used a metric learning loss as an auxiliary term to improve the performance.
Differently, our PDML framework models pixels as probabilistic distributions and then computes the distance between distributions to address the spectral uncertainty and label uncertainty.
\section{Proposed Approach}
In this section, we first formulate the problem of hyperspectral image classification and provide a unified view of existing spectral-spatial-based hyperspectral image classification methods. Then, we present the uncertainty modeling method of spectral distribution. After that, we introduce how to measure the distances between distributions and how to conduct the discriminative loss to instruct the training procedure. Lastly, we elaborate on the proposed probabilistic deep metric learning framework for HSIC.

\subsection{Revisit Spectral–Spatial-Based HSIC }
Consider a hyperspectral image set $\mathbf{X}$ composed of $N$ training samples $\{\mathbf{x}_1,\mathbf{x}_2,\cdots,\mathbf{x}_N\}$, in which each sample $\mathbf{x}_i$ contains $M$ pixels $\{\mathbf{p}_i^1,\mathbf{p}_i^2,\cdots,\mathbf{p}_i^M\}$ with their corresponding high dimensional spectral vectors $\mathbf{S}_i = \{s_i^1,s_i^2,\cdots,s_i^M\}$ and ground truth labels $\mathbf{L}_i = \{l_i^1, l_i^2, \cdots , l_i^M \}$, where $l_i^m \in \{1,2,\cdots,K\}$ indicates that $\mathbf{p}_i^m$ belongs to the $l_i^m$th class. Let $c_i^m$ be the real appearance of the objects in the pixel $\mathbf{p}_i^m$. General HSIC tasks aim at learning an accurate mapping from the real appearance of the objects in the pixel $\mathbf{p}_i^m$ to the predicted label of the pixel $y_i^m$, which can be regarded as predicting a probability $p(y_i^m|c_i^m)$. As for spectral-feature networks for HSIC, they only consider the spectral information of a given pixel and predict a relative probability $p(y_i^m|s_i^m)$. In most previous studies, the probability was computed by a softmax function, which is formulated as follows:
\begin{eqnarray}\label{equ:softmax}
p(y_i^m=k|s_i^m) = \frac{\exp({V_k})}{\sum_{j=1}^{K}\exp({V_j})},
\end{eqnarray}
where $V_k$ denotes the $k$th number in the output vector of the deep network. Obviously, the softmax function normalizes the numbers in the output vector and maps them to a range from $0$ to $1$. Besides, the softmax function is always followed by a cross entropy loss which optimizes the whole network for better classification performance:
\begin{eqnarray}\label{equ:cross entropy}
L = -\sum_{k=1}^{K}{\mathbb{I}(l_i^m=k)log(p(y_i^m=k|s_i^m))},
\end{eqnarray}
where $\mathbb{I}(\cdot)$ denotes the indicator function which takes $1$ as the output when the expression evaluates to true while outputs $0$ otherwise.

However, the aforementioned spectral variability demonstrates that even the same captured object may present different spectral features, which means the probability $p(s_i^m|c_i^m)$
exists. Under such circumstances, the procedure of the spectral-based deep learning methods could be formulated as follows:
\begin{eqnarray}\label{equ:spectral}
p(y_i^m|s_i^m) = H(p(y_i^m|c_i^m),p(s_i^m|c_i^m)),
\end{eqnarray}
where $H(\cdot)$ represents a non-linear mapping. Therefore, only using the spectral information of a hyperspectral image to perform hyperspectral classification tasks might be influenced by the spectral variability and result in unfaithful performance. 

Given the above consideration, spatial information has been involved in HSIC in order to weaken the spectral uncertainty and improve the classification performance. To be specific, spatial-spectral-based methods tend to utilize the neighbor information of a pixel and take a patch as input. That is to say, a $5\times5$ patch includes the central pixel and the other $24$ pixels around it. In addition, the label of the patch is absolutely the same as that of the central pixel and the subsequent processing is also conducted by the softmax function and the cross entropy loss. For a better understanding of spatial-spectral-based methods, we provide a simple probabilistic explanation of why these methods truly work. 

Low as the resolution of hyperspectral images is, the labels of the neighbor pixels are usually the same as the center pixel, which signifies that the pixels in a patch tend to encode the same material. Thus, we suppose that obtaining the wrong prediction of a certain category of pixels is of the same probability $p_0$. On such assumption, the accuracy of classifying a single hyperspectral pixel using the spectral-based methods is $1-p_0$. However, as for the spatial-spectral-based approaches, the corresponding probability becomes closer to $1-p_0^T$, where $T$ denotes the number of pixels in a patch ($T=25$ when the patch size is $5\times5$, for instance). In view of $p_0<1$ in most cases, we can easily conclude that $1-p_0^T > 1-p_0$, which indicates that the classification accuracy of spatial-spectral-based methods is bound to surpass the accuracy of spectral-based approaches. Moreover, multiscale hyperspectral methods simultaneously consider multiple patches of diverse sizes, which utilizes more spatial information and further enhances the classification performance.

Nevertheless, the existence of low spatial resolution of hyperspectral images results in the problem that different pixels and even the same pixel in a patch are likely to encode diverse materials (a pixel contains both water and land, for example). Therefore, the aforementioned probability $p_0$ should not be the same for each pixel in a patch all the time and simply attaching the label of the central pixel to the neighbor pixels would sometimes ignore abundant spatial information and cause serious problems. For instance, provided a pixel is at the edge of a ground object, the patch center on this pixel will beyond all doubt contain a certain number of pixels that are definitely of different categories compared with the central one. On such occasions, forcing the deep model to classify the above patch according to the label of the central pixel sounds unreasonable. In other words, the labels of the patches are to some extent of uncertainty and previous spatial-spectral-based hyperspectral classification approaches have not taken the label uncertainty into consideration.

\begin{figure*}[t]
\centering
\includegraphics[width=0.96\textwidth]{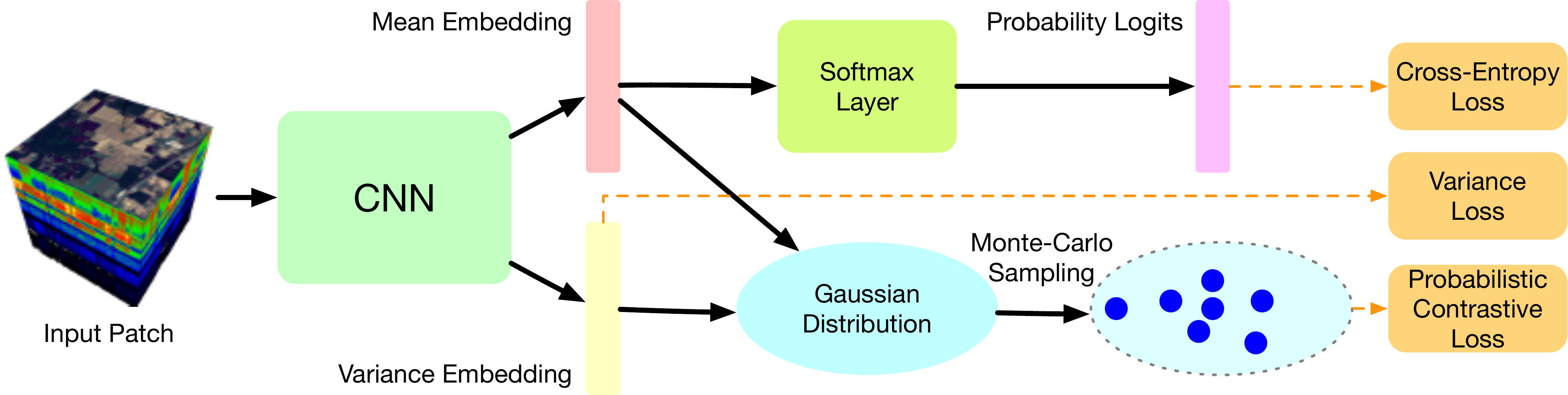}
\caption{An illustration of the proposed PDML framework. 
We first employ a backbone CNN to extract meaningful features from the input hyperspectral patch.
We then obtain the mean and variance embeddings of each pixel in the patch and model each pixel as a high-dimensional Gaussian distribution.
We only demonstrate one pixel in the figure for simplicity. 
We employ Monte-Carlo sampling to generate a few samples from the Gaussian distribution and use them to compute the match probability of a pair of pixels.
We then impose the probabilistic contrastive loss to enlarge the interclass match probabilities and decrease the intraclass match probabilities.
We further use a variance loss to constrain the variance of an outer pixel to be larger than that of an inner pixel. 
We further employ a softmax layer and the cross-entropy loss to train a classifier to predict the category of the center pixel.
The overall framework can be trained simultaneously in an end-to-end manner.
(Best viewed in color.)
}
\label{fig:framework}
\vspace{-5mm}
\end{figure*}

\subsection{Uncertainty Modeling of Spectral Distribution}
In order to effectively extract more information from the hyperspectral patches and achieve better classification performance, we attempt to directly model the label uncertainty by introducing the spectral distribution of each pixel in a patch. The modeling operation mainly consists of two separate parts: the basic convolutional neural network structure and the uncertainty modeling part.

Specifically, the convolutional neural network aims at extracting abundant features from the input hyperspectral patches, which takes the advantage of the local perception of each input figure and is widely used in computer vision tasks from diverse research fields. In our research, let $\mathbf{D}$ be the patch set and $\{\mathbf{d}_1,\mathbf{d}_2,\cdots,\mathbf{d}_N\}$ be the corresponding patch samples with the size of $s\times s\times d$, where $s\times s$ denotes the patch size while $d$ denotes the spectral dimension of the original images. As mentioned above, the ground truth label of each patch is the same as the label of the central pixel in the patch. Therefore, the processing of the CNN architecture can be formulated as $\mathbf{f}_i = \mathbf{g}(\mathbf{d}_i)$, where $\mathbf{f}_i$ denotes the output feature of the convolutional network of the $i$th input. 
We further control the patch size of the output features to be consistent with the input patches, which is of benefit to the subsequent study. 
On the contrary, the size of the spectral dimension decreases through each layer of the CNN to avoid the redundancy of the original high-dimensional spectral information.

With the corresponding extracted features of the same patch size as the input patches, we utilize the Gaussian distribution to model the categorical uncertainty of the spectral distribution. Concretely, we consider each pixel in a patch and model each pixel as a high-dimensional Gaussian distribution but not a point embedding in previous research. To obtain the corresponding mean and variance embeddings of each pixel, we employ two convolutional layers which map the extracted features into a high-dimensional embedding space in which the mean matrix and the variance matrix are of the same size $s\times s\times r$, where $r$ denotes the dimension of the Gaussian distribution of each pixel. The above mapping can be formulated as follows ($T=s\times s$ denoting the number of pixels in a certain patch):
\begin{eqnarray}
&&\mathbf{M}_i = \mathbf{g}_1(\mathbf{f}_i) = \{m_i^1,m_i^2,...,m_i^T\}, \label{equ:Mean} \\
&&\mathbf{V}_i = \mathbf{g}_2(\mathbf{f}_i) = \{v_i^1,v_i^2,...,v_i^T\}, \label{equ:variance}
\end{eqnarray}
where the mean matrix $\mathbf{M}_i$ and the variance matrix $\mathbf{V}_i$ can be regarded as the combination of $T$ vectors with the size of $r$. Besides, the $k$th vector in $\mathbf{M}_i$ ($m_i^k$) and the corresponding $k$th vector in $\mathbf{V}_i$ ($v_i^k$) respectively represent the mean and variance parameters of the Gaussian distribution of the $k$th pixel in the $i$th patch.
After the above treatment, each pixel in the patch has been mapped to a corresponding stochastic embedding. In addition, we assign all of the embeddings in a patch with the label of the center pixel, which makes it possible for the subsequent training procedure.

\subsection{Deep Probabilistic Metric Learning}
Having obtained the Gaussian distribution as well as the label information of each pixel embedding, we attempt to optimize the parameters of these distributions and effectively utilize the distributions to enhance the overall classification performance of our architecture.

Firstly, in view of the fact that outer pixels in a patch tend to encode more different materials with the central pixel compared with inner pixels, indicating that outer pixels possess more label uncertainty, we enforce a larger Gaussian variance for outer pixels, which can be achieved by controlling the average variance of outer pixels to be larger than that of inner pixels by a predefined margin. Specifically, a $s\times s$ patch contains $\frac{s+1}{2}$ lap pixels and the $j$th lap contains $n_j$ pixels, which can be formulated as follows:
\begin{eqnarray}
n_j=
\begin{cases}
1 &  j=1, \\
(2j-1)^2-(2j-3)^2 & j\geq 2.
\end{cases}
\end{eqnarray}
Therefore, the constraint on the variance of the $j$th lap is:
\begin{eqnarray}\label{equ:variance control}
\mathbf{J}_{var(j)}^{(i)} = -(\frac{1}{n_j}\sum_{k}v_i^k - (1+\alpha)\frac{1}{n_{j-1}}\sum_{q}v_i^q),
\end{eqnarray}
where $j=\{2,3,...,\frac{s+1}{2}\}$, $\alpha$ is a predefined sensitive parameter to control the variance differences between neighboring laps of pixels, $\sum_{k}v_i^k$ denotes adding the variance of the pixel embeddings from the $j$th lap and $\sum_{q}v_i^q$ denotes the same procedure from the $(j-1)$th lap.

Furthermore, we directly obtain multiple samples from the aforementioned Gaussian distributions via the Monte-Carlo sampling method and we utilize a probabilistic metric learning loss function to model the corresponding distance between them. For example, suppose two pixels are mapped to the embedding space and the Gaussian distributions of them are $p(\mathbf{z}_1|\mathbf{p}_1) \sim N(m_1,v_1)$ and $p(\mathbf{z}_2|\mathbf{p}_2) \sim N(m_2,v_2)$ respectively. Therefore, the Monte-Carlo sampling strategy can be recognized as follows:
\begin{eqnarray}
z_1^{(k_1)} = m_1 + v_1\cdot \epsilon^{(k_1)}, \label{equ:Monte-Carlo} \\
z_2^{(k_2)} = m_2 + v_2\cdot \epsilon^{(k_2)}, \label{equ:Monte-Carlo} 
\end{eqnarray}
where $k_1=\{1,2,...,K\}$ and $k_2=\{1,2,...,K\}$ indicate extracting $K$ individual samples from each Gaussian distribution, $\epsilon^{(k_1)} ,\epsilon^{(k_2)} \sim N(0,\mathbf{I})$. With the above samples and the corresponding ground truth label information, we can utilize various deep metric learning losses to conduct the optimization, such as contrastive loss and triplet loss. Differently, in our research, we employ a probabilistic contrastive loss to enlarge the interclass distances as well as reduce the intraclass distances of the samples. 
Specifically, we introduce the conventional contrastive loss function and the probabilistic contrastive loss formation, and we also detail how the probabilistic contrastive loss can be used in our Gaussian samples.

The original contrastive loss is trained on a pair of training data $(z_i,z_j)$ together with a sign parameter $y_{ij}$ which is $1$ when the two samples are from the same class and is $0$ otherwise. The contrastive loss directly minimizes the distances between the positive pairs and punishes the distances between the negative pairs for being larger than a predefined margin $\beta$, which is formulated as follows:
\begin{eqnarray}\label{equ:contrastive loss}
J_{con} = \frac{1}{B}\sum_{(i,j)}^{B/2}y_{ij}D_{ij}^2+(1-y_{ij})[\beta-D_{ij}]_+^2,
\end{eqnarray}
where $B$ denotes the batch size of the training samples,  $[\cdot]_+=max(\cdot,0)$, and $D_{ij}=D(z_i,z_j)=||z_i-z_j||_2$. 

However, the conventional contrastive loss is absolutely a certain formation of metric learning losses because the relations between two samples merely include match and non-match. Thus, it is necessary to provide a probabilistic contrastive loss formation that can represent the probability that two samples are matching and is more appropriate for our probabilistic embedding strategy. The match probability of two samples is defined as follows:
\begin{eqnarray}\label{equ:match probability}
p(z_i,z_j)=\sigma(-aD_{ij}+b),
\end{eqnarray}
where $\sigma(\cdot)=\frac{1}{1+e^{-(\cdot)}}$ is the sigmoid function varying from $0$ to $1$, while $a>0$ and $b\in R$ are two learnable parameters standing for the soft threshold of the pairwise distances. Under such circumstances, the match probability ranges from $0$ to $1$ and increases when the distance $D_{ij}$ decreases, which is consistent with our objective.

With the match probability of two samples $p(z_i,z_j)$, the probabilistic contrastive loss is defined as:
\begin{eqnarray}\label{equ:probabilistic contrastive loss}
\mathbf{J}_{pcon}(z_i,z_j)=
\begin{cases}
-\log p(z_i,z_j) & \mathbf{if\quad match} ,\\
-\log (1-p(z_i,z_j)) & \mathbf{otherwise},
\end{cases}
\end{eqnarray}
where the probabilistic contrastive loss enlarges the match probability when the two samples are from the same class while reducing the probability for negative pairs.

As for the samples extracted from the Monte-Carlo sampling, we directly add the pairwise match probabilities of the various samples from two Gaussian distributions and consider the result as the match probability of the two distributions, which can be formulated as follows:
\begin{eqnarray}
p(\mathbf{p}_1,\mathbf{p}_2) \approx \frac{1}{K^2} \sum_{k_1=1}^K \sum_{k_2=1}^K p(z_1^{(k_1)},z_2^{(k_2)}),
\end{eqnarray}
where we use the pixel pair $(\mathbf{p}_1,\mathbf{p}_2)$ to represent the Gaussian distributions of pixels in the embedding space for simplicity.

Therefore, we apply the probabilistic contrastive loss to the match probability of distributions:
\begin{eqnarray}\
\mathbf{J}_{pcon}(\mathbf{p}_1,\mathbf{p}_2)=
\begin{cases}
-\log p(\mathbf{p}_1,\mathbf{p}_2) & \mathbf{if\quad match} ,\\
-\log (1-p(\mathbf{p}_1,\mathbf{p}_2)) & \mathbf{otherwise}.
\end{cases}
\end{eqnarray}

Finally, we combine the constraint on the variance of the Gaussian distributions and the probabilistic contrastive loss for a batch of input patches, which can be formulated as follows:
\begin{eqnarray}
\mathbf{J}_{dist} = \lambda_1\frac{1}{B}(\sum_{i=1}^B \sum_{j=2}^{(s+1)/2}\mathbf{J}_{var(j)}^{(i)}) \nonumber\\
+ \lambda_2\frac{1}{B\times T}(\sum_{i,j}\mathbf{J}_{pcon}(\mathbf{p}_i,\mathbf{p}_j)),
\end{eqnarray}
where $\lambda_1$ and $\lambda_2$ are pre-defined parameters to balance the contributions of different losses while $T=s\times s$ denotes the number of pixels in a patch, $\sum_{i,j}\mathbf{J}_{pcon}(\mathbf{p}_i,\mathbf{p}_j)$ indicates adding all of the probabilistic contrastive loss terms of the pairwise distributions from the embedding space in the batch.

Our probabilistic deep metric learning framework not only models the label uncertainty of the input hyperspectral patches but also takes advantage of more samples in the embedding space for training, which can exploit full information from each input patch.

\renewcommand{\algorithmicrequire}{\textbf{Input:}}
\renewcommand{\algorithmicensure}{\textbf{Output:}}
\setlength{\textfloatsep}{7pt}
\begin{algorithm}[tb]
\caption{\!\!\! \textbf{:} PDML} \label{alg: PDML}
\begin{algorithmic}[1]
\Require
Training image set, labels, learning rates, iteration number $T$, network weights $\theta$, variance controlling parameter $\alpha$, and sampling number of the Monte-Carlos sampling $K$.
\Ensure
Parameters $\theta$, and classification results.
\For{$iter=1,2,\cdots,T$}
\State Utilize the CNN modules to extract features $\textbf{f}$ from the input images.
\State Obtain the mean matrix and the variance matrix of each pixel through two convolution layers respectively.
\State Generate K samples from the Gaussian distribution using the Monte-Carlo sampling strategy.
\State Employ a probabilistic metric learning loss to restrain the distances of the samples.
\State Control the Gaussian variance of outer pixels to be larger than that of inner pixels and compute the corresponding loss.
\State Map the mean matrix to get the classification results with a softmax structure and apply the cross entropy loss to compute the classification errors.
\State Update the network parameters $\theta$ with the above losses.
\EndFor\\
\Return
$\theta$.
\end{algorithmic}
\end{algorithm}

\subsection{Hyperspectral Image Classification with PDML}
To conduct hyperspectral image classification with our probabilistic metric learning framework, we involve the aforementioned probabilistic constraint in the conventional hyperspectral classification methods. Specifically, we employ a convolutional layer on the mean matrix $\mathbf{M}_i$ followed by a softmax structure to obtain the classification results, which can be formulated as follows:
\begin{eqnarray}
p(y_i=k)=\frac{\exp(f_l(\mathbf{M}_i)_k)}{\sum_{j=1}^K\exp(f_l(\mathbf{M}_i)_j)},
\end{eqnarray}
where $y_i$ denotes the predicted label of the $i$th patch, $f_l(\cdot)$ denotes the non-linear mapping, and $f_l(\mathbf{M}_i)_k$ denotes the $k$th term of $f_l(\mathbf{M}_i)$. Therefore, the cross entropy loss term is:
\begin{eqnarray}
\mathbf{J}_{ce}=-\sum_{i}\sum_{k=1}^{K}{\mathbb{I}(l_i=k)log(p(y_i=k))},
\end{eqnarray}
where $l_i$ denotes the ground truth label of the $i$th hyperspectral patch and $\mathbf{J}_{ce}$ indicates adding all the cross entropy terms of the patches in an input batch. In summary, the objective function of the whole PDML framework to conduct HSIC is:
\begin{eqnarray}
\mathbf{J}_{PDML}=\mathbf{J}_{dist} + \lambda_3 \mathbf{J}_{ce},
\end{eqnarray}
where $\lambda_3$ is a balance factor. 

During training, the parameters of the layers which map the mean matrix to the classification results will only be updated by the cross entropy term while the convolutional layer mapping the features to the variance matrix will merely be optimized by the Gaussian probabilistic loss. Nevertheless, other parts of our architecture will be optimized by both loss function terms. The overall framework of our proposed PDML is illustrated in Figure~\ref{fig:framework} and the full training procedure of our method can be found in Algorithm~\ref{alg: PDML}.

During testing, we only pass the hyperspectral patches through the mean matrix network and then use the softmax function to obtain the probability that the input patches belong to each category, which introduces no additional workload compared with conventional methods.

It is worth mentioning that we can apply our framework to various existing hyperspectral image classification approaches. First of all, the feature extracted architecture can be replaced by numerous structures, such as diverse 3D-CNN, 2D-CNN, and the recurrent neural network, which might respectively acquire different features of the input hyperspectral image. In addition, with the generated samples in the embedding space, the probabilistic contrastive loss function can be substituted with popular metric learning functions like margin loss with distance weighted sampling strategy, which could further effectively select valuable samples to optimize the network for better performance. Besides, the widely used attention modules in computer vision tasks could also be added to our architecture, which focuses on more discriminative positions and channels of the hyperspectral images and is beneficial to the classification task.

\newcommand{\tablesize}{\small}

\begin{table*}[t] \tablesize
\centering
\caption{Classification Results of different methods for the IN dataset.}
\vspace{-1mm}
\setlength\tabcolsep{4pt}
\renewcommand\arraystretch{1.1}
\begin{tabular}{cccccccccc}
\toprule
Class & SVM & SAE & EMAP & CNN & SPC & MCNN & 3D-CNN & SSRN & PDML \\
\midrule 
OA(100\%) & 81.67$\pm$0.65 & 85.47$\pm$0.58 & 91.65$\pm$0.63 & 97.41$\pm$0.43 & 90.68$\pm$0.75 & 98.30$\pm$0.20 & 97.56$\pm$0.43 & 99.19$\pm$0.26 & \textbf{99.55$\pm$0.09}\\
AA(100\%) & 79.84$\pm$3.37 & 86.34$\pm$1.14 & 95.15$\pm$1.05 & 97.39$\pm$0.56 & 92.00$\pm$2.84 & 97.40$\pm$0.40 &  99.23$\pm$0.19 & 98.93$\pm$0.59 & \textbf{99.37$\pm$0.21}\\
$\kappa \times 100$ & 78.76$\pm$0.77 & 83.42$\pm$0.66 & 90.46$\pm$0.68 & 83.42$\pm$0.66 & 89.36$\pm$0.86 & 98.00$\pm$0.30 & 97.02$\pm$0.52 & 99.07$\pm$0.30 & \textbf{99.48$\pm$0.12}\\
\midrule 
1 & 96.78 & 81.82 & 94.87 & \textbf{100.00} & 83.15 & 98.30 & \textbf{100.00} & 97.82 & 99.46\\
2 & 78.74 & 82.16 & 84.10 & 97.27 & 86.81 & 96.20 & 96.34 & 99.17 & \textbf{99.66}\\
3 & 82.26 & 77.54 & 95.66 & 98.00 & 87.34 & 95.30 & 99.49 & 99.53 & \textbf{100.00}\\
4 & 99.03 & 68.11 & 98.37 & 92.81 & 91.32 & 94.20 & \textbf{100.00} & 97.79 & \textbf{100.00}\\
5 & 93.75 & 94.36 & 95.53 & 99.25 & 97.54 & 98.20 & \textbf{99.91} & 99.24 & 99.23\\ 
6 & 85.96 & 94.45 & 96.84 & 99.52 & 97.88 & 98.70 & \textbf{99.75} & 99.51 & 99.65\\ 
7 & 40.00 & 94.70 & \textbf{100.00} & 97.58 & 89.33 & 96.40 & \textbf{100.00} & 98.70 & \textbf{100.00}\\ 
8 & 91.80 & 94.36 & 99.32 & 99.00 & 90.85 & \textbf{100.00} & \textbf{100.00} & 99.85 & 98.93\\ 
9 & 0.00 & 82.56 & \textbf{100.00} & 96.95 & \textbf{100.00} & 92.20 & \textbf{100.00} & 98.50 & 98.22\\ 
10 & 86.00 & 81.28 & 90.63 & 95.38 & 81.92 & 97.80 & 98.72 & 98.74 & \textbf{99.15}\\ 
11 & 70.94 & 84.47 & 89.16 & 97.72 & 91.68 & 99.60 & 95.52 & 99.30 & \textbf{99.65}\\ 
12 & 74.73 & 83.77 & 86.88 & 97.13 & 85.14 & 97.20 & 99.47 & 98.43 & \textbf{99.56}\\ 
13 & 99.04 & 96.42 & 98.77 & 99.65 & 99.72 & \textbf{100.00} & \textbf{100.00} & \textbf{100.00} & \textbf{100.00}\\ 
14 & 94.29 & 92.27 & 94.13 & 97.95 & 97.44 & 99.70 & 99.55 & 99.31 & \textbf{99.79}\\ 
15 & 85.11 & 80.63 & 98.18 & 92.30 & 93.43 & 98.80 & \textbf{99.54} & 99.20 & 98.69\\ 
16 & 96.78 & 81.82 & \textbf{100.00} & \textbf{100.00} & 83.15 & 95.80 & 99.34 & 97.82 & 97.86\\
\bottomrule
\end{tabular}
\label{tab:IN_result}
\vspace{-2mm}
\end{table*}

\begin{table*}[tb] \tablesize
\centering
\caption{Classification Results of different methods for the UP dataset.}
\vspace{-1mm}
\setlength\tabcolsep{4pt}
\renewcommand\arraystretch{1.1}
\begin{tabular}{cccccccccc}
\toprule
Class & SVM & SAE & EMAP & CNN & SPC & MCNN & 3D-CNN & SSRN & PDML \\
\midrule 
OA(100\%) & 90.58$\pm$0.47 & 94.25$\pm$0.18 & 93.52$\pm$0.34 & 98.85$\pm$0.15 & 98.88$\pm$0.22 & 99.50$\pm$0.20 & 99.54$\pm$0.11 & 99.79$\pm$0.09 & \textbf{99.91$\pm$0.03} \\
AA(100\%) & 92.99$\pm$0.36 & 93.34$\pm$0.39 & 94.82$\pm$0.40 & 98.40$\pm$0.30 & 98.40$\pm$0.27 & 99.30$\pm$0.10 & 99.66$\pm$0.11 & 99.66$\pm$0.17 & \textbf{99.89$\pm$0.07}\\
$\kappa \times 100$ & 87.21$\pm$0.70 & 92.35$\pm$0.25 & 91.65$\pm$0.49 & 98.47$\pm$0.20 & 98.52$\pm$0.30 & 99.30$\pm$0.20 & 99.41$\pm$0.15 & 99.72$\pm$0.12 & \textbf{99.82$\pm$0.09}\\
\midrule
1 & 87.24 & 94.59 & 91.30 & 98.98 & 99.01 & 99.60 & 99.36 & 99.92 & \textbf{99.98}\\
2 & 89.93 & 96.44 & 91.83 & 99.45 & 99.81 & 99.40 & 99.36 & \textbf{99.96} & 99.94\\
3 & 86.48 & 84.57 & 72.22 & 96.40 & 95.46 & 99.40 & \textbf{99.69} & 98.46 & 99.68\\
4 & 99.95 & 97.37 & 99.80 & 99.58 & 99.54 & 99.80 & 99.63 & 99.69 & \textbf{100.00}\\
5 & 95.78 & 99.60 & 99.93 & 99.39 & 99.84 & 99.90 & 99.95 & \textbf{99.99} & \textbf{99.99}\\ 
6 & 97.69 & 99.39 & 99.62 & 99.70 & 99.18 & 99.80 & 99.96 & 99.94 & \textbf{99.98}\\ 
7 & 95.44 & 88.57 & 99.70 & 97.18 & 98.15 & 98.80 & \textbf{100.00} & 99.82 & 99.89\\ 
8 & 84.40 & 85.66 & 99.27 & 95.73 & 94.65 & 98.90 & \textbf{99.65} & 99.22 & 99.55\\ 
9 & \textbf{100.00} & 99.88 & 99.79 & 99.56 & 99.99 & 98.40 & 99.38 & 99.95 & \textbf{100.00}\\ 
\bottomrule
\end{tabular}
\label{tab:UP_result}
\vspace{-4mm}
\end{table*}

\begin{table*}[tb] \tablesize
\centering
\caption{Classification Results of different methods for the KSC dataset.}
\vspace{-1mm}
\setlength\tabcolsep{4pt}
s\begin{tabular}{cccccccccc}
\toprule
Class & SVM & SAE & EMAP & CNN & SPC & MCNN & 3D-CNN & SSRN & PDML \\
\midrule 
OA(100\%) & 80.20$\pm$0.58 & 92.99$\pm$0.59 & 91.23$\pm$0.59 & 97.08$\pm$0.47 & 97.90$\pm$0.49 & 98.49$\pm$0.65 & 96.30$\pm$1.25 & 99.61$\pm$0.22 & \textbf{99.71$\pm$0.18}\\
AA(100\%) & 65.64$\pm$0.86 & 89.76$\pm$1.25 & 89.17$\pm$1.64 & 95.09$\pm$0.70 & 96.56$\pm$0.69 & 97.12$\pm$0.62 & 94.68$\pm$1.97 & 99.33$\pm$0.57 & \textbf{99.44$\pm$0.31}\\
$\kappa \times 100$ & 77.97$\pm$0.65 & 92.18$\pm$0.91 & 90.25$\pm$0.54 & 96.74$\pm$0.53 & 97.66$\pm$0.55 & 98.32$\pm$0.74 & 95.90$\pm$1.39 & 99.56$\pm$0.25 & \textbf{99.67$\pm$0.18}\\
\midrule 
1 & 92.16 & 93.04 & 94.09 & 99.00 & 99.11 & 99.44 & 91.71 & 99.70 & \textbf{99.96}\\
2 & 86.16 & 92.04 & 86.15 & 98.48 & 99.19 & 98.82 & 89.73 & 99.88 & 99.81\\
3 & 42.55 & 85.59 & 98.05 & 92.16 & 92.60 & \textbf{100.00} & 92.16 & 99.00 & 99.08\\
4 & 67.69 & 72.12 & 67.33 & 81.84 & 85.49 & 82.95 & 86.94 & 98.26 & \textbf{98.58}\\
5 & 0.00 & 82.20 & 71.32 & 85.38 & 89.63 & 84.07 & 94.79 & 99.03 & 96.67\\ 
6 & 54.71 & 83.15 & 89.13 & 90.96 & 95.94 & \textbf{100.00} & 90.92 & 99.43 & 99.68\\ 
7 & 0.00 & 76.46 & 95.24 & 93.21 & 96.38 & 97.30 & 91.57 & 97.03 & \textbf{99.34}\\ 
8 & 65.12 & 94.10 & 88.70 & 98.21 & 98.09 & \textbf{100.00} & 96.22 & 99.54 & 99.61\\ 
9 & 67.82 & 94.57 & 95.43 & 99.04 & 99.53 & \textbf{100.00} & 99.53 & 99.70 & 99.97\\ 
10 & 93.40 & 98.91 & \textbf{100.00} & 99.85 & 99.96 & \textbf{100.00} & 99.81 & 99.96 & \textbf{100.00}\\ 
11 & \textbf{100.00} & 98.39 & 93.45 & 98.89 & 99.86 & \textbf{100.00} & 97.69 & 99.80 & \textbf{100.00}\\ 
12 & 83.75 & 96.42 & 84.12 & 99.43 & 99.51 & \textbf{100.00} & 97.69 & \textbf{100.00} & 99.97\\ 
13 & \textbf{100.00} & 99.83 & 96.23 & 99.79 & 99.97 & \textbf{100.00} & \textbf{100.00} & \textbf{100.00} & \textbf{100.00}\\ 
\bottomrule
\end{tabular}
\label{tab:KSC_result}
\vspace{-2mm}
\end{table*}

\begin{table*}[tb] \tablesize
\centering
\caption{Classification Results of different methods for the Houston 2013 dataset.}
\vspace{-1mm}
\setlength\tabcolsep{8.7pt}
\renewcommand\arraystretch{1.1}
\begin{tabular}{ccccccccccc}
\toprule
Class & CNN & ECNN & GCNN & 3D-CNN & MSDNSA & SSRN & CASSN & RIAN & SSAtt  & PDML\\
\midrule 
OA(100\%) & 61.85 & 84.04 & 84.12 & 78.19 & 87.78 & 89.46 & 85.32 & 86.37 & 90.38 &  \textbf{92.62$\pm$0.47}\\
AA(100\%) & 62.98 & 83.33 & 82.94 & 75.79 & 89.28 & 89.45 & 87.28 & 88.68 & 89.76 & \textbf{93.89$\pm$0.81}\\
$\kappa \times 100$ & 58.64 & 82.54 & 82.51 & 76.27 & 86.73 & 88.58 & 84.07 & 85.14 & 89.55 & \textbf{91.99$\pm$0.56}\\
\midrule 
1 & 56.73 & 87.49 & 87.47 & 78.63 & 82.72 & 81.48 & 83.10 & 83.00 & 82.54 & \textbf{96.30}\\
2 & 64.68 & 80.99 & 86.01 & 93.23 & 99.81 & 92.48 & 85.15 & 83.74 & \textbf{99.92} & 99.62\\
3 & 44.67 & 87.72 & 78.22 & 40.99 & 89.70 & 98.02 & \textbf{100.00} & 89.70 & 86.48 & \textbf{100.00}\\
4 & 59.05 & 90.43 & 85.02 & 97.44 & 95.08 & 98.11 & 93.28 & 92.52 & \textbf{99.57} & 97.54\\
5 & 68.31 & \textbf{100.00} & 99.89 & 87.31 & 94.89 & 99.91 & 99.24 & 99.81 & 99.61 & 99.72\\ 
6 & 70.21 & 97.90 & 89.44 & 79.02 & 95.80 & 95.80 & 95.80 & \textbf{100.00} & 83.71 & 99.30\\ 
7 & 82.57 & 90.48 & 90.19 & \textbf{90.49} & 85.63 & 89.46 & 82.65 & 87.59 & 89.92 & \textbf{90.49}\\ 
8 & 52.12 & 58.51 & 74.44 & 59.83 & 85.57 & 69.90 & 66.48 & 73.22 & 81.94 & \textbf{90.12}\\ 
9 & 70.35 & 79.77 & 84.42 & 81.11 & \textbf{86.02} & 84.04 & 79.60 & 81.21 & 85.99 & 85.36\\ 
10 & 60.37 & 64.28 & 63.61 & 69.59 & 60.33 & 82.34 & 66.80 & 68.15 & \textbf{88.81} & 87.64\\ 
11 & 72.16 & 78.37 & 80.06 & 75.14 & 87.67 & \textbf{93.17} & 90.42 & 89.85 & 90.53 & 81.21\\ 
12 & 44.63 & 78.29 & 87.30 & 82.23 & 90.78 & \textbf{90.80} & 89.15 & 89.15 & 89.48 & 88.76\\ 
13 & 87.02 & 76.84 & 85.06 & 82.11 & 90.88 & 72.98 & 77.54 & 92.28 & 86.81 & \textbf{92.98}\\ 
14 & 96.92 & 99.19 & \textbf{100.00} & 80.57 & 99.60 & 99.19 & \textbf{100.00} & \textbf{100.00} & 95.34 & \textbf{100.00}\\ 
15 & 14.93 & 77.04 & 56.95 & 39.11 & 94.71 & 94.08 & \textbf{100.00} & \textbf{100.00} & 85.77 & 99.37\\ 
\bottomrule
\end{tabular}
\label{tab:Hou_result}
\vspace{-4mm}
\end{table*}

\section{Experiments}
In this section, we conducted various experiments to evaluate the classification performance of the proposed PDML framework on four widely-used benchmark datasets: Indian Pines (IN) dataset, University of Pavia (UP) dataset, Kennedy Space Center (KSC) dataset, and Houston 2013 dataset.

\subsection{Datasets}
For fair comparisons with existing methods, we randomly partitioned the IN, UP, and KSC datasets into the training, validation, and testing subsets following~\cite{article}, while we constructed the training and testing subsets for the Houston 2013 dataset following~\cite{debes2014hyperspectral}. 
We provide the detailed information of each dataset as follows:

\begin{itemize}

\item The \textbf{IN} dataset was collected by the Airborne Visible Infrared Imaging Spectrometer (AVIRIS) in Northwest Indiana in 1996. It contains 16 different land-cover classes of $145\times 145$ pixels. Originally, the Indian Pines dataset has 220 spectral channels with a resolution of 20 m by pixel. Since 20 bands are corrupted by water absorption effects, we discard the above 20 bands and utilize the remaining 200 bands for research, which cover the range from 0.4 to 2.5$\mu m$. Specifically, we randomly choose 20$\%$, 10$\%$, and 70$\%$ of the labeled data for training, validation, and testing respectively. 
 
\item The \textbf{UP} dataset was gathered by the Reflective Optics System Imaging Spectrometer in Italy in 2001, which contains 9 land-cover classes of 610 $\times$ 340 pixels from the city of Pavia. After the noisy bands are removed from the hyperspectral images, the remaining images include 103 spectral bands with a resolution of 1.3 m by pixel and range from 0.43 to 0.86$\mu m$. Different from the IN dataset, we randomly divide the UP dataset into training, validation, and testing subsets while the ratio of the division is 10$\%$, 10$\%$, and 80$\%$, respectively.

\item The \textbf{KSC} dataset was also gathered by AVIRIS from the Kennedy Space Center located in Florida in 1996, which contains 13 types of swamps of 512 $\times$ 614 pixels with a spatial resolution of 18 m by pixel. In addition, we obtain relative data with 176 spectral bands without the water absorption bands. Similar to the IN dataset, we randomly divide the KSC dataset into training, validation, and testing subsets and the ratio of the division is the same as that of the IN dataset.

\item The \textbf{Houston 2013} dataset was collected from the University of Houston campus in 2012~\cite{debes2014hyperspectral}. The dataset contains 15 different land covers with 144 spectral bands and a spatial size of 349 $\times$ 1905. 
We partitioned the dataset into training and testing subsets following~\cite{debes2014hyperspectral}.

\end{itemize}

\subsection{Implementation Details}
We employed the Tensorflow framework to conduct all the experiments. 
We adopted part of the architecture in SSRN~\cite{article} as the trunk CNN model and added 2 randomly initialized convolutional layers as the embedding structure to obtain the corresponding mean and variance embedding.
We standardized all the hyperspectral images of the four datasets to zero mean value and unit variance value.
We fixed the batch size to 16 for the IN, UP, and KSC datasets and 128 for the Houston 2013 dataset.
We used the RMSProp optimizer~\cite{tieleman2012lecture} with learning rates of $10^{-4}$ for all the datasets for fair comparisons.
Besides, we set the input patch size to $5\times5$, the sensitive parameter $\alpha$ to $0.2$, and the number $K$ of our Monte-Carlo sampling to $3$ in our experiments.
The parameter $\lambda_1$, $\lambda_2$, and $\lambda_3$ were set to 1.
We ran $300$ epochs during the training procedure and utilized the model which performed the best in the validation samples to test the classification results for the IN, UP, and KSC datasets. For the Houston 2013 dataset, we provided the best testing performance of the whole training procedure.
We repeated all the experiments ten times and reported the average classification results including the overall accuracy (OA), the average accuracy (AA), the Kappa coefficient, and the per-class accuracy.

\begin{table}[t]\scriptsize
\caption{OA (\%) and $\kappa$ of PDML on various architectures for the UP dataset.}
\centering
\vspace{-1mm}
\setlength\tabcolsep{1pt}
\renewcommand\arraystretch{1.1}
\begin{tabular}{c|cccccc}
\hline
Model & HNet & P-HNet & LNet & P-LNet & DFFN & P-DFFN\\
\hline 
OA(100\%) & 93.10$\pm$0.96 & \textbf{94.89$\pm$0.89} & 96.61$\pm$0.91 & \textbf{97.80$\pm$0.67} & 97.99$\pm$0.37 & \textbf{98.81$\pm$0.27}\\
$\kappa \times 100$ & 90.92$\pm$1.26 & \textbf{93.34$\pm$1.05} & 95.63$\pm$0.99 & \textbf{97.13$\pm$0.63} & 97.05$\pm0.42$ & \textbf{98.56$\pm$0.32}\\
\hline
\end{tabular}
\label{tab:implementation}
\vspace{-2mm}
\end{table}

\begin{table}[t] \tablesize
\centering
\caption{OA (\%) of PDML with different loss functions in the embedding space.}
\vspace{-1mm}
\setlength\tabcolsep{2pt}
\renewcommand\arraystretch{1.1}
\begin{tabular}{c|cccc}
\hline
Loss & IN & UP & KSC & Houston 2013\\
\hline 
P. Contrastive & \textbf{99.55$\pm$0.09} & \textbf{99.91$\pm$0.03} & \textbf{99.71$\pm$0.18} & \textbf{92.62$\pm$0.47}\\
Contrastive & 99.21$\pm$0.17 & 99.82$\pm$0.08 & 99.58$\pm$0.18 & 91.88$\pm$0.62\\
Triplet-R  & 98.45$\pm$0.41 & 99.19$\pm$0.13 & 99.04$\pm$0.46 & 91.34$\pm$0.65\\
Triplet-SH & 98.52$\pm$0.33 & 99.32$\pm$0.08 & 98.87$\pm$0.42 & 92.29$\pm$0.51\\
Npair & 98.63$\pm$0.71 & 99.11$\pm$0.25 & 98.94$\pm$0.51 & 90.75$\pm$0.73\\
Softmax & 98.55$\pm$0.68 & 99.17$\pm$0.33 & 98.67$\pm$0.49 & 90.32$\pm$0.82\\
\hline
\end{tabular}
\label{tab:loss}
\vspace{-2mm}
\end{table}

\subsection{Results and Analysis}
\textbf{Comparisions with state-of-the-art methods:}
We compared the proposed PDML framework with state-of-the-art hyperspectral image classification methods including
SVM~\cite{waske2010sensitivity}, SAE~\cite{chen2014deep}, EMAP~\cite{dalla2010morphological}, CNN~\cite{makantasis2015deep}, SPC~\cite{article}, MCNN~\cite{li2019convolutional}, 3D-CNN~\cite{chen2016deep}, ECNN~\cite{aptoula2016deep}, GCNN~\cite{chen2017hyperspectral}, MSDNSA~\cite{fang2019hyperspectral}, SSAtt~\cite{hang2020hyperspectral}, CASSN~\cite{9641863},
RIAN~\cite{9785505}, and SSRN~\cite{article}.
Tables~\ref{tab:IN_result}, ~\ref{tab:UP_result}, ~\ref{tab:KSC_result}, and ~\ref{tab:Hou_result} show the quantitative experimental results on the widely used IN, UP, KSC, and Houston 2013 datasets, respectively.
We observe that our PDML obtained the best classification performance in all four datasets. For instance, the mean value of the overall accuracy (OA) of our method for the UP dataset is the highest while the standard deviation value is the lowest among all of the compared approaches. Additionally, for the challenging Houston 2013 dataset, our PDML framework achieves significantly better results than existing methods. This is because our method exploits more information in the input patches and directly models the label uncertainty of the neighboring pixels in a probabilistic form.

\newcommand\figwidth{0.228}
\begin{figure}[t]
\centering
\subcaptionbox{ 1\% training samples\label{subfig:1}}{
\includegraphics[width=\figwidth\textwidth]{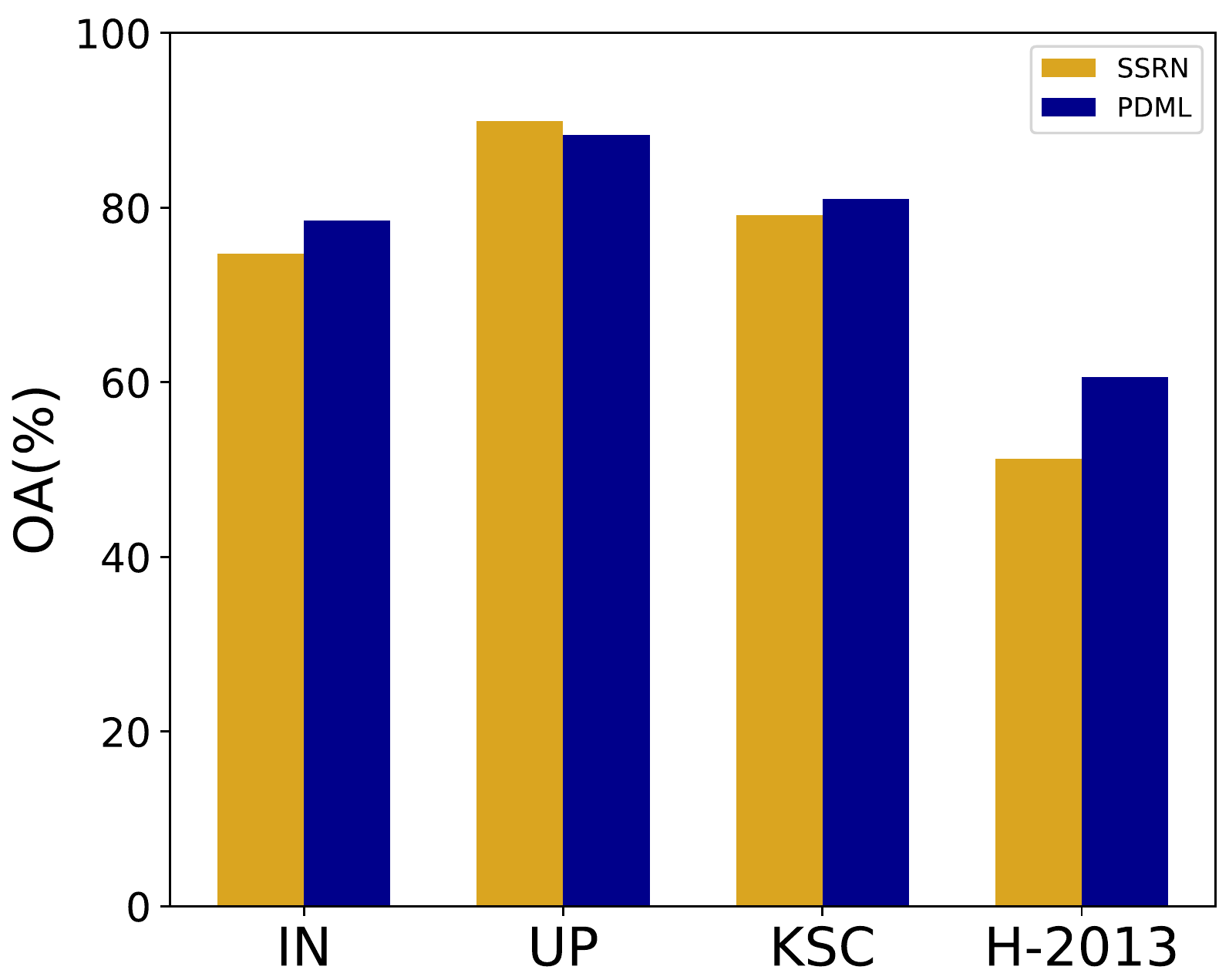}
}\hfill
\subcaptionbox{ 5\% training samples\label{subfig:5}}{
\includegraphics[width=\figwidth\textwidth]{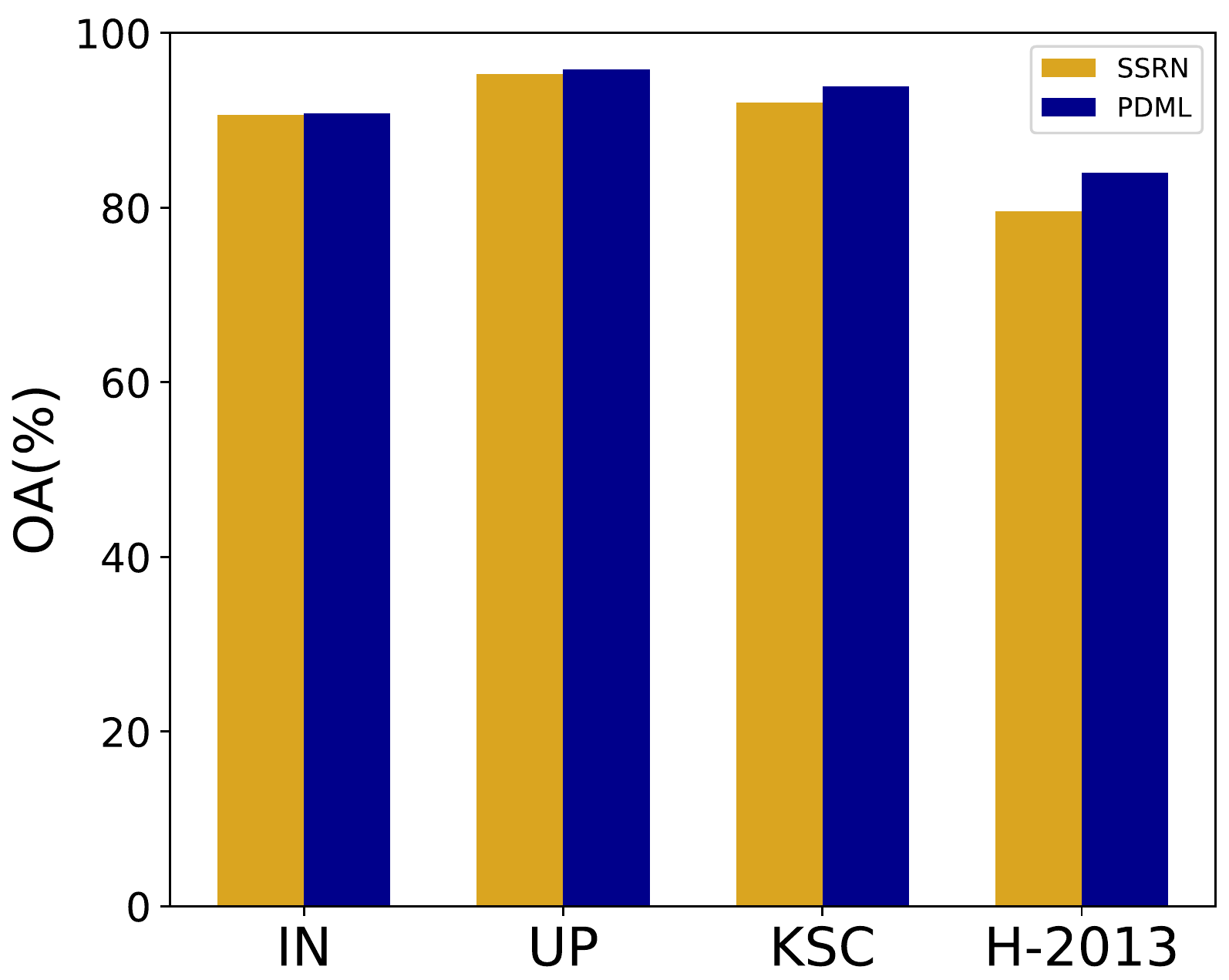}
}
\caption{The experimental results with fewer training samples on four datasets using SSRN and PDML.}
\label{fig:sample}
 \vspace{-4mm}
\end{figure}

\newcommand\figwidthh{0.228}
\begin{figure}[t]
\centering
\subcaptionbox{ Training sample proportions.\label{fig:training_proportions}}{
\includegraphics[width=\figwidthh\textwidth]{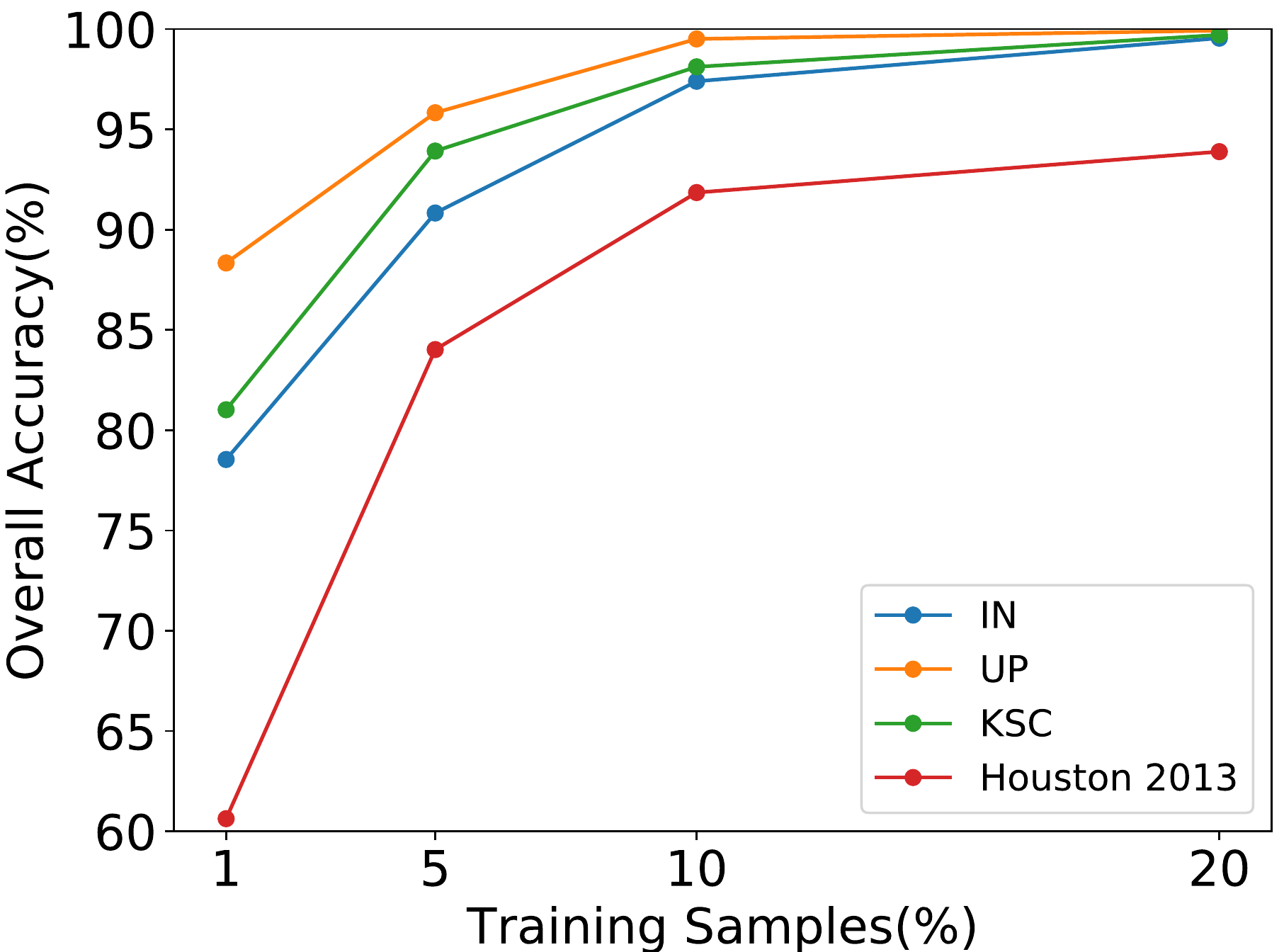}
}\hfill
\subcaptionbox{ Patch size.\label{fig:patch_size}}{
\includegraphics[width=\figwidthh\textwidth]{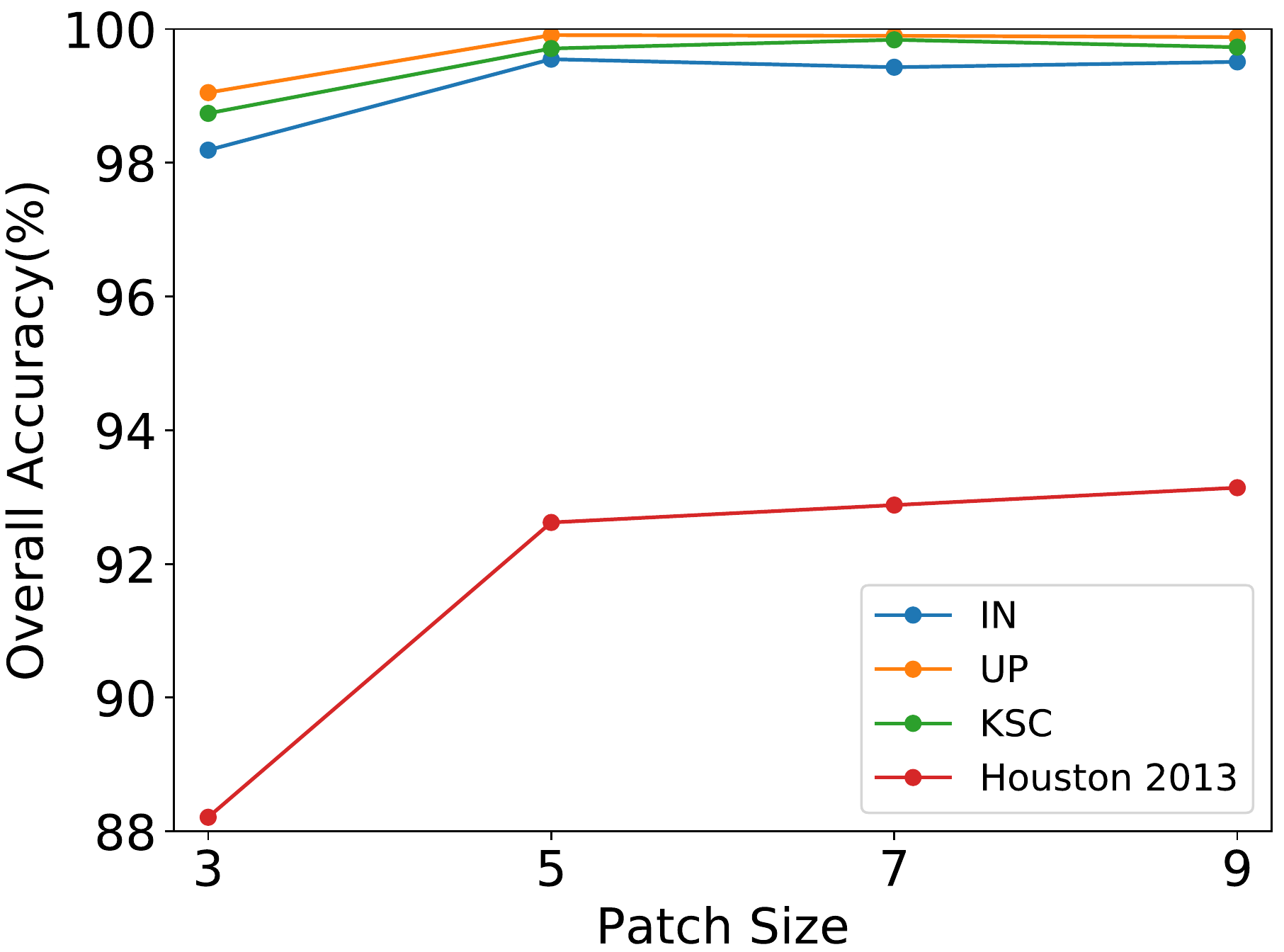}
}
\caption{OA (\%) of PDML with different factors in the IN, UP, KSC, and Houston 2013 datasets.}
\label{fig:tsne}
\vspace{-2mm}
\end{figure}

\begin{table}[t]\caption{OA (\%) of PDML with different numbers of Gaussian distributions for IN, UP, KSC, and Houston 2013.}
\centering
\vspace{-1mm}
\setlength\tabcolsep{6pt}
\renewcommand\arraystretch{1.1}
\begin{tabular}{c|cccc}
\hline
Number & IN & UP & KSC & Houston 2013\\
\hline 
1 & \textbf{99.55$\pm$0.09} & \textbf{99.91$\pm$0.03} & 99.71$\pm$0.18 & 92.62$\pm$0.47\\
2 & 99.36$\pm$0.09 & 99.88$\pm$0.05 & 99.68$\pm$0.15 & \textbf{92.91$\pm$0.53}\\
3 & 99.01$\pm$0.13 & 99.65$\pm$0.06 & \textbf{99.74$\pm$0.16} & 91.69$\pm$0.66\\
\hline
\end{tabular}
\label{tab:Gaussian}
\vspace{-2mm}
\end{table}

\begin{table}[t]\caption{OA (\%) of PDML with different $\lambda_1$, $\lambda_2$, and $\lambda_3$ on four datasets.}
\centering
\vspace{-1mm}
\setlength\tabcolsep{4pt}
\renewcommand\arraystretch{1.1}
\begin{tabular}{ccc|cccc}
\hline
$\lambda_1$ & $\lambda_2$ & $\lambda_3$ & IN & UP & KSC & Houston 2013\\
\hline 
1.0 & 1.0 & 1.0 & 99.55$\pm$0.09 & 99.91$\pm$0.03 & 99.71$\pm$0.18 & 92.62$\pm$0.47\\
1.0 & 1.0 & 0.5 & 99.48$\pm$0.09 & 99.89$\pm$0.04 & 99.65$\pm$0.22 & 92.48$\pm$0.55\\
1.0 & 1.0 & 0.0 & 72.55$\pm$1.88 & 78.49$\pm$1.27 & 70.91$\pm$2.43 & 46.73$\pm$2.24\\
1.0 & 0.5 & 1.0 & 99.45$\pm$0.15 & 99.89$\pm$0.05 & 99.63$\pm$0.21 & 91.89$\pm$0.71\\
1.0 & 0.0 & 1.0 & 99.23$\pm$0.16 & 99.81$\pm$0.05 & 99.61$\pm$0.25 & 89.85$\pm$0.65\\
0.5 & 1.0 & 1.0 & 99.51$\pm$0.11 & 99.91$\pm$0.03 & 99.66$\pm$0.16 & 92.59$\pm$0.42\\
0.0 & 1.0 & 1.0 & 99.49$\pm$0.09 & 99.89$\pm$0.04 & 99.65$\pm$0.18 & 92.23$\pm$0.46\\
\hline
\end{tabular}
\label{tab:parameter}
\vspace{-2mm}
\end{table}

\begin{table}[t]\caption{The computation time (seconds) of one epoch for PDML compared with the SSRN baseline for the IN, UP, KSC and Houston 2013 datasets. 
}
\centering
\vspace{-1mm}
\setlength\tabcolsep{7pt}
\renewcommand\arraystretch{1.1}
\begin{tabular}{c|cccc}
\hline
Method & IN & UP & KSC & Houston 2013\\
\hline 
SSRN (Training) & 262.69 & 1006.74 & 136.41 & 151.68\\
PDML (Training) & 986.61 & 3621.60 & 532.67 & 693.71\\
\hline
SSRN (Testing) & 32.18 & 156.29 & 17.03 & 59.58\\
PDML (Testing) & 31.95 & 161.21 & 17.42 & 57.49\\
\hline
\end{tabular}
\label{tab:time}
\vspace{-2mm}
\end{table}

\textbf{Application of PDML to existing methods:}
We applied our PDML framework to other deep models to test the effectiveness of the method. Specifically, the models include HamidaNet~\cite{hamida20183} (HNet), LiNet~\cite{li2017spectral} (LNet), and DFFN~\cite{song2018hyperspectral}. 
As for the structure of these models, HNet and LNet are both based on the 3-D CNN blocks while the number of layers and the size of the CNN layer vary with each other. Differently, DFFN is based on 2-D CNN block.
We constructed the corresponding architectures of the models to test the original classification performance and then replaced the last fully connected layer with two convolutional layers to generate the embedding mean matrix and the variance matrix as we have described before, followed by the same probabilistic procedure.
Table~\ref{tab:implementation} shows the original classification results of the three models that we have obtained by our implementation together with the results after we added the PDML framework to the original architectures. We observe a constant performance boost to the models on both the overall accuracy and the kappa coefficient. For example, our PDML framework delivered a $1.79\%$ increase in the overall accuracy and a $2.42\%$ increase in the kappa coefficient for LNet, which dramatically proves that our proposed framework can be properly implemented on various deep models and comprehensively enhances the classification results of the original architectures.

\begin{figure*}[t]
\centering
\includegraphics[width=0.9\textwidth]{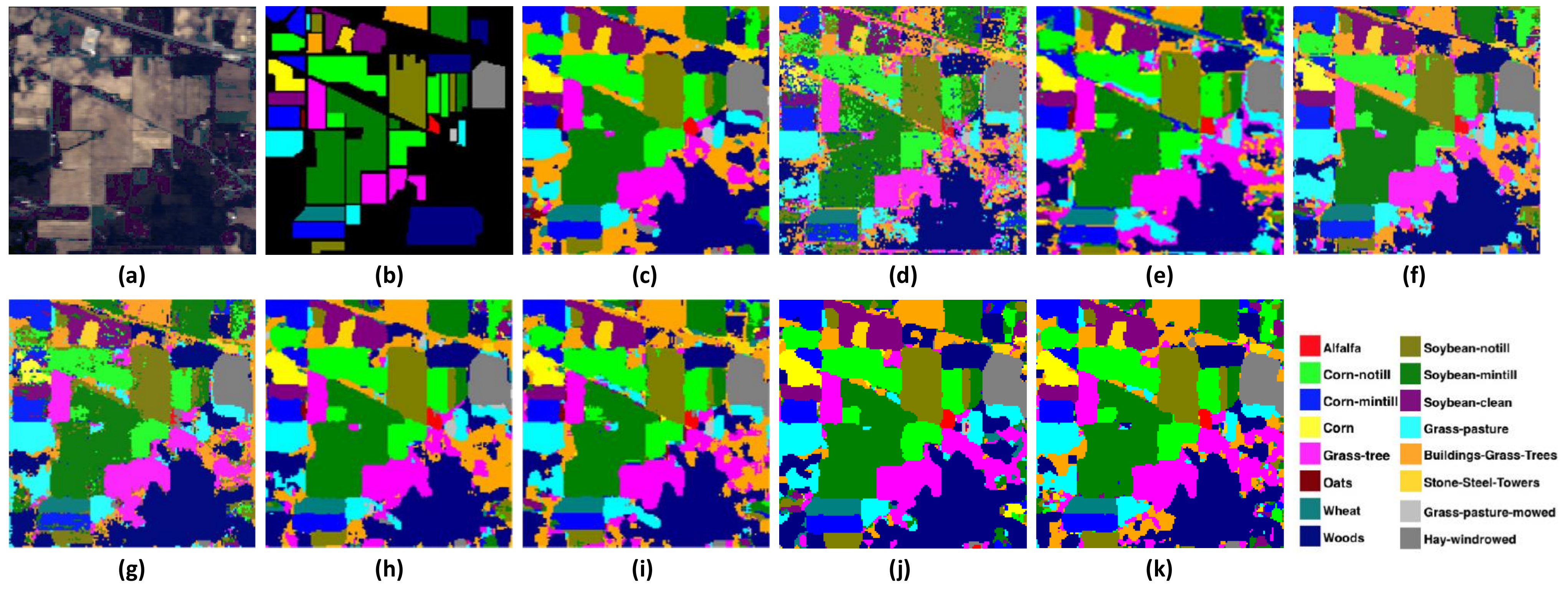}
\vspace{-2mm}
\caption{Classification maps for the Indian Pines dataset. (a) False-color image. (b) Ground-truth map. (c)-(k) Classification maps of SVM, SAE, EMAP, CNN, SPC, MCNN, 3D-CNN, SSRN, and PDML.
(Best viewed in color.)
}
\label{fig:IN_classification_map}
\vspace{-5mm}
\end{figure*}

\begin{figure*}[t]
\centering
\includegraphics[width=0.9\textwidth]{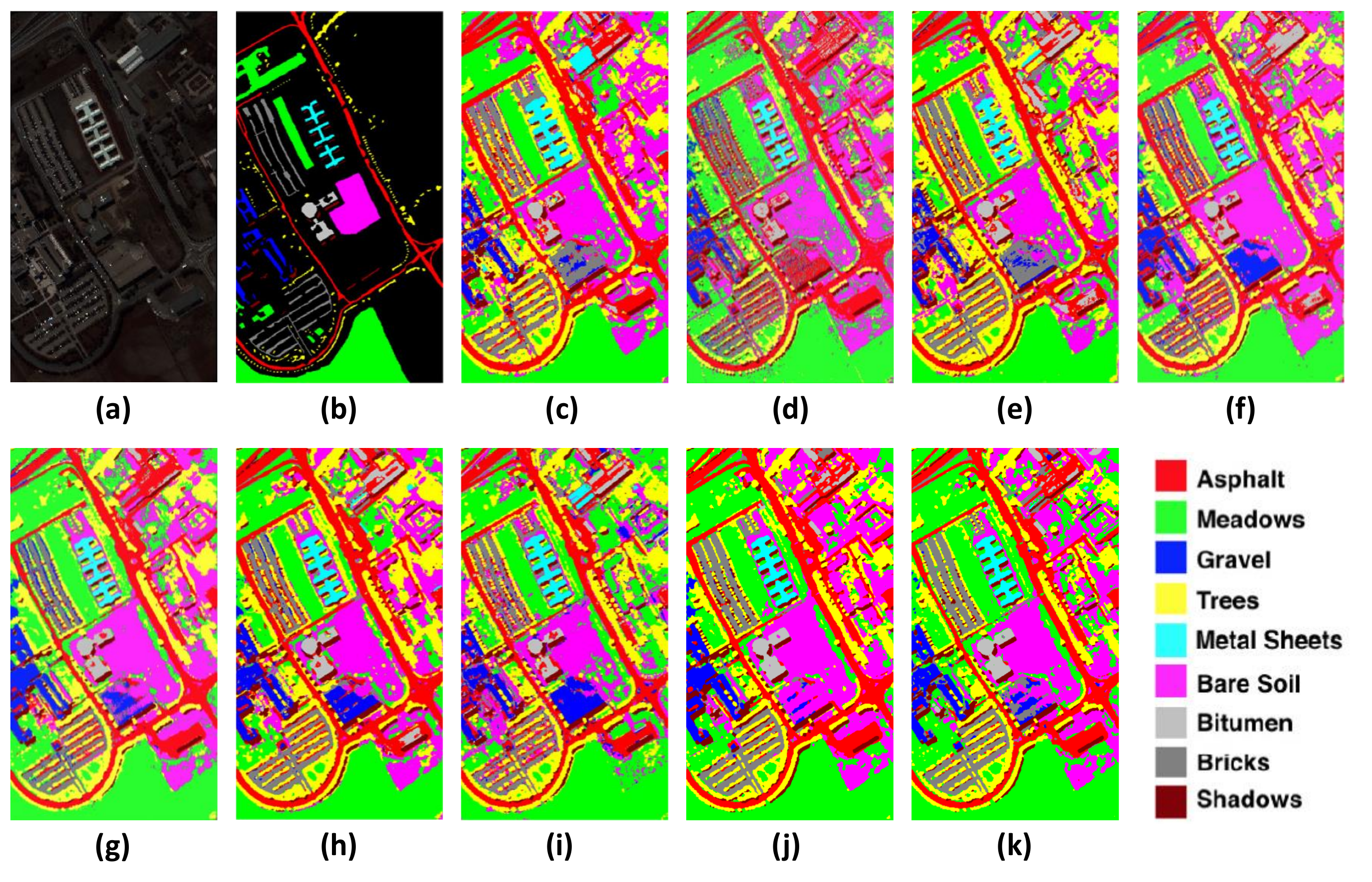}
\vspace{-3mm}
\caption{Classification maps for the University of Pavia dataset. (a) False-color image. (b) Ground-truth map. (c)-(k) Classification maps of SVM, SAE, EMAP, CNN, SPC, MCNN, 3D-CNN, SSRN, and PDML.
(Best viewed in color.)
}
\label{fig:UP_classification_map}
\vspace{-5mm}
\end{figure*}

\begin{figure*}[t]
\centering
\includegraphics[width=0.9\textwidth]{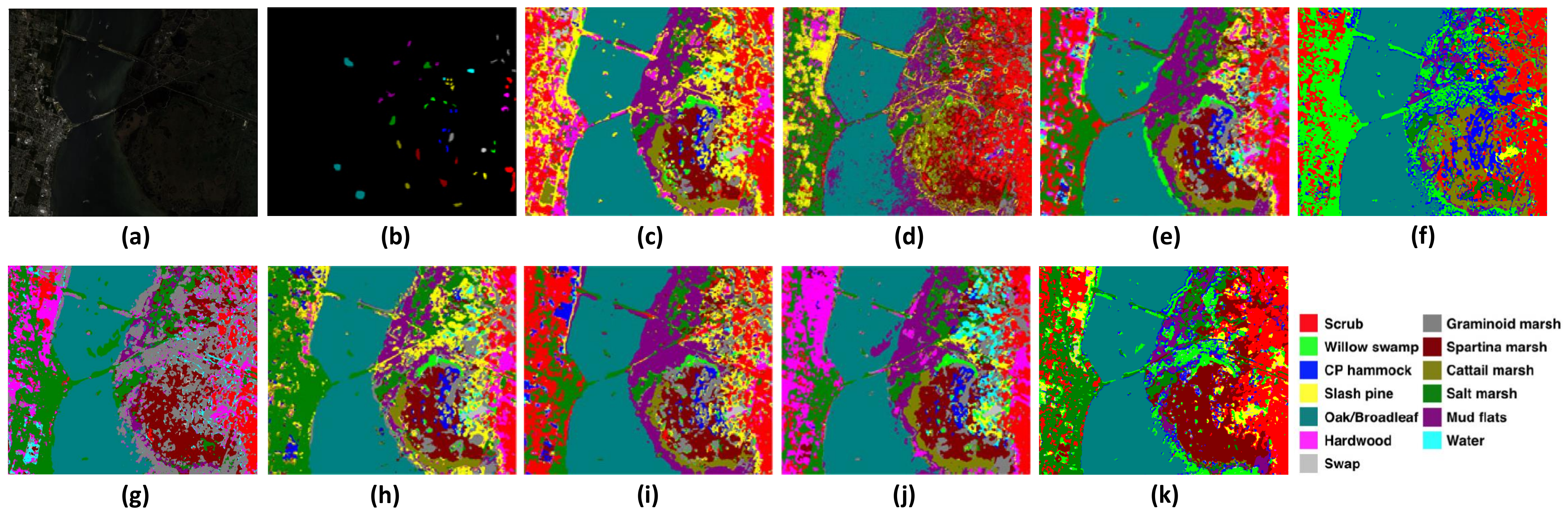}
\caption{Classification maps for the Kennedy Space Center dataset. (a) False-color image. (b) Ground-truth map. (c)-(k) Classification maps of SVM, SAE, EMAP, CNN, SPC, MCNN, 3D-CNN, SSRN, and PDML.
(Best viewed in color.)
}
\label{fig:KSC_classification_map}
\vspace{-5mm}
\end{figure*}

\begin{figure*}[t]
\centering
\includegraphics[width=0.9\textwidth]{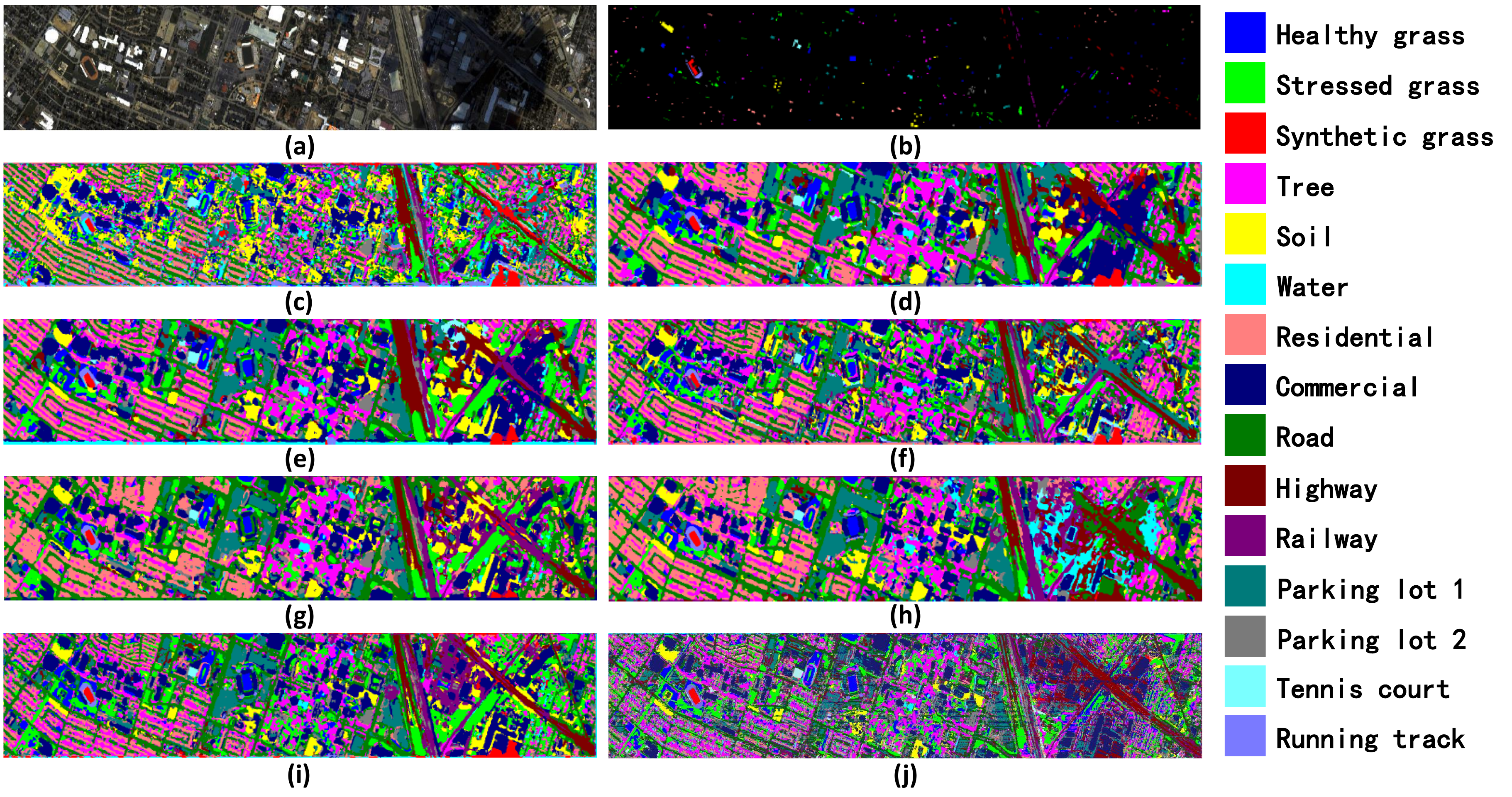}
\caption{Classification maps for the Houston 2013 dataset. (a) False-color image. (b) Ground-truth map. (c)-(j) Classification maps of CNN, ECNN, GCNN, 3D-CNN, MSDNSA, SSRN, SSAtt, and PDML.
(Best viewed in color.)
}
\label{fig:Hou_classification_map}
\vspace{-5mm}
\end{figure*}

\begin{figure}[t]
\centering
\subcaptionbox{Before training.\label{subfig:before}}{
\includegraphics[width=0.21\textwidth]{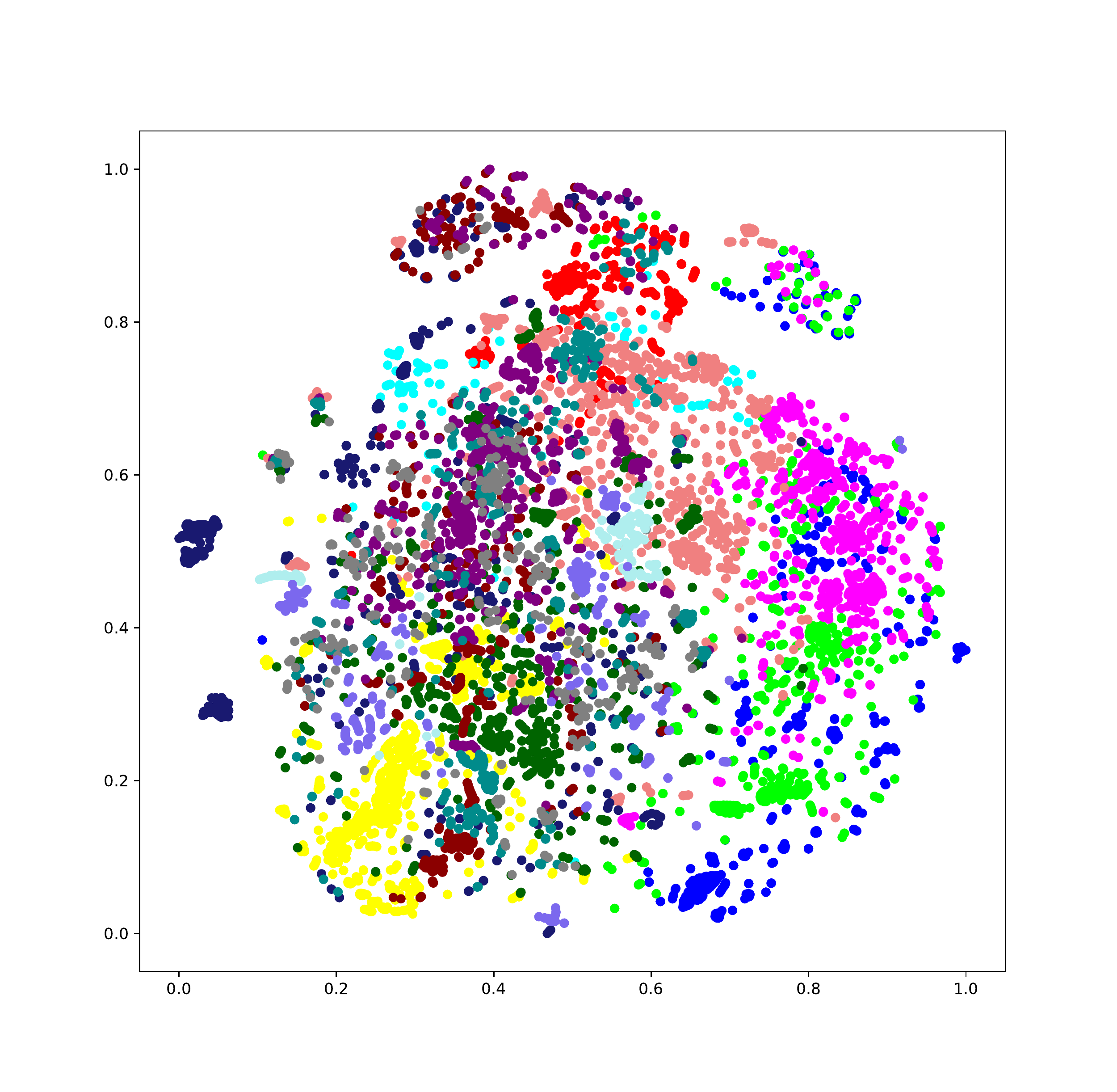}
}\hfill
\subcaptionbox{After training.\label{subfig:after}}{
\includegraphics[width=0.21\textwidth]{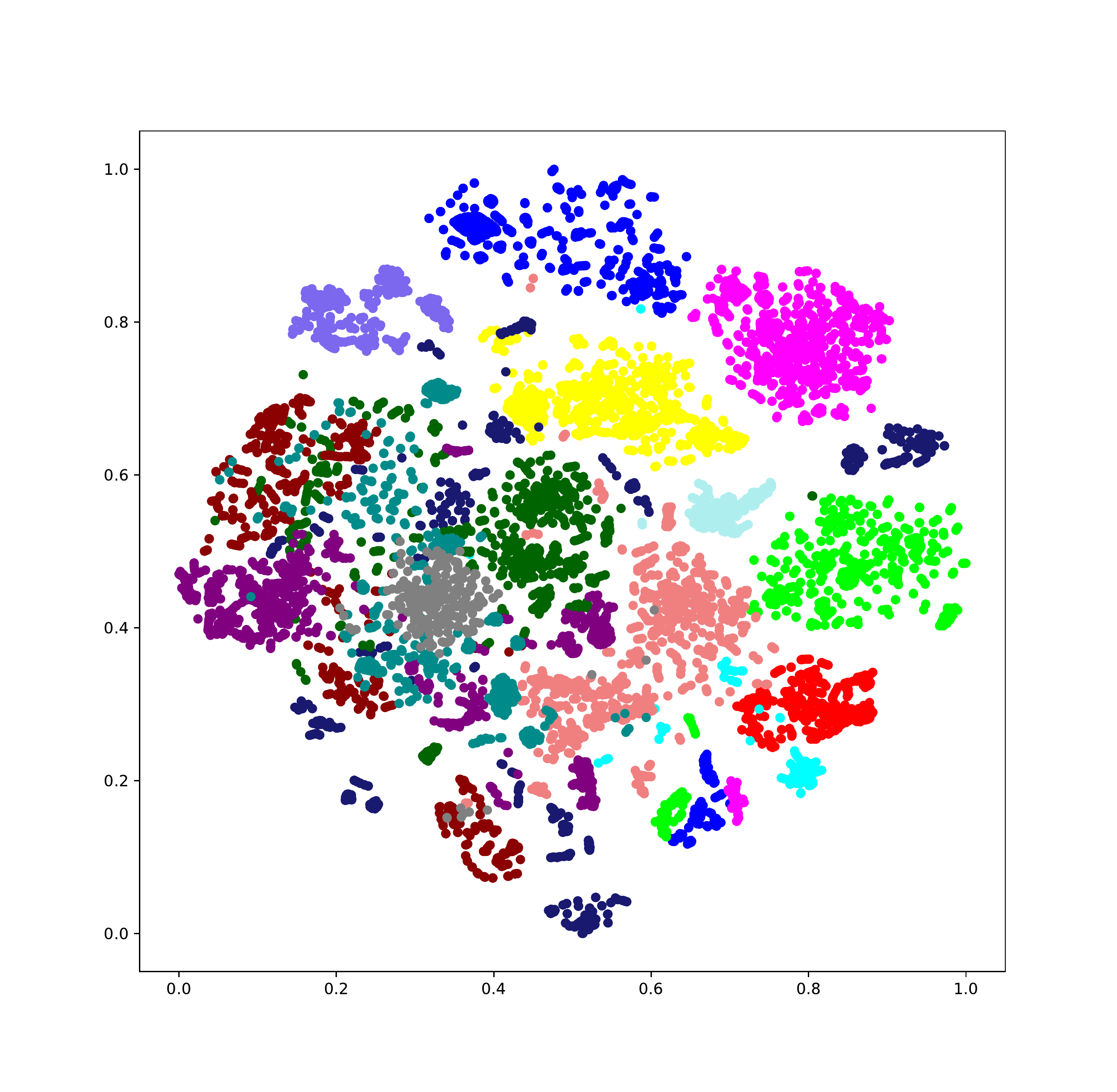}
}
\caption{The t-SNE plots of the embeddings for the Houston 2013 dataset before and after training. (Best viewed in color.)}
\label{fig:tsne}
\vspace{-2mm}
\end{figure}

\textbf{Effect of the probabilistic contrastive loss:}
We conducted a series of ablation experiments on four datasets to analyze the effect of the probabilistic contrastive loss in the proposed PDML framework.
To be specific, we respectively substitute the probabilistic contrastive loss with the conventional contrastive loss, the triplet loss with random sampling, the triplet loss with semi-hard sampling, the N-pair loss, and the softmax loss while keeping other structures unchanged during training. 

The results are shown in Table~\ref{tab:loss}. We can see that the probabilistic contrastive loss outperforms other loss structures in all four datasets. That is because the probabilistic contrastive loss introduces the match probability of two samples which offers a better judgment of samples as well as involves more learnable parameters which are more reasonable and flexible for deep models.
By comparison, the contrastive loss and the triplet-based losses behave better than the N-pair loss and the softmax loss.
This is because the contrastive loss and the triplet loss are more effective for the clustering in the embedding space, which might be more suitable in our settings.

\textbf{Effect of the training sample proportions:}
We compared the performances of SSRN and PDML with fewer training samples on four datasets, as shown in Figure~\ref{fig:sample}. We observe that PDML achieves consistent performance boost compared with SSRN with 1\% and 5\% training samples.
In addition, we respectively used 1\%, 5\%, 10\%, and 20\% of labeled pixels in the datasets to train our PDML framework. The experimental results are shown in Figure~\ref{fig:training_proportions}. We see that the overall accuracy increases when taking more training samples for all the datasets. Additionally, the classification results are competitive as the proportions reach 10\%, and using more training samples does not significantly boost the performance.

\textbf{Effect of the patch sizes:}
The patch size influences the spatial information involved for each pixel. Therefore, we considered using the patch size of $3\times3$, $5\times5$, $7\times7$, and $9\times9$ to conduct the experiment. As illustrated in Figure~\ref{fig:patch_size}, the IN, UP, and KSC datasets are less sensitive to the patch size and we observe the best performance when the patch size is set to $5\times5$. As for the Houston 2013 dataset, the overall accuracy markedly increases as the patch size increases from $3\times3$ to $5\times5$. Nevertheless, the classification results indicate little enhancement for larger patch size but involve extra computational cost on the contrary. Under such circumstances, we fix the patch size to $5\times5$ for subsequent experiments.

\textbf{Effect of the forms of Gaussian distributions:}
In the proposed PDML framework, we model each pixel with a single Gaussian distribution to consider the data uncertainty. We further verified the performances of PDML when applying multiple Gaussian distributions. We present the classification results in Table~\ref{tab:Gaussian}. We see that modeling each pixel with multiple Gaussian distributions is not beneficial to the performance. Furthermore, the overall accuracy even decreases when using 3 Gaussian distributions for the Houston 2013 dataset, which demonstrates that a single Gaussian distribution is capable of modeling each pixel, and utilizing multiple Gaussian distributions might involve negative influences.

\textbf{Effect of $\lambda_1$, $\lambda_2$, and $\lambda_3$:}
The parameter $\lambda_1$, $\lambda_2$, and $\lambda_3$ were set to 1 during training, which demonstrates an equal weight for three loss functions. We respectively fixed two parameters and modified the value of the other parameter to analyze the contribution of each parameter. Table~\ref{tab:parameter} presents the experimental results. Firstly, the overall accuracy significantly decreases when $\lambda_3$ is set to 0 because the final classification layer can not be properly trained but provides random outputs. Additionally, we see a worse performance when we reduce the weight of $\lambda_2$, which certifies that our probabilistic deep metric learning term indeed boosts the overall classification results. Particularly, when $\lambda_2=0$, our framework is similar to the conventional hyperspectral image classification framework, which demonstrates the universality of our method. Moreover, we observe that $\lambda_1$ influences the overall accuracy to some extent, demonstrating the effectiveness of enforcing a large variance for outer pixels in a patch.

\textbf{Computation time:}
We provide the computation time of one epoch for PDML and SSRN~\cite{article} in Table~\ref{tab:time}. 
Note that we use a larger batch size for the Houston 2013 dataset, resulting in higher efficiency. 
We see that PDML requires more computation time during training on four datasets compared with SSRN due to the Monte-Carlo sampling.
However, our method introduces no additional computation workload during inference.
Therefore, it is fair to directly compare our PDML with the baseline methods.
We think the increase in the training time is acceptable given the superior performance and similar inference time of our framework.

\textbf{Visualization of classification results:}
We qualitatively visualize the classification maps of the best trained models with their corresponding false-color images and ground truth maps for four datasets in Figure~\ref{fig:IN_classification_map}, ~\ref{fig:UP_classification_map}, ~\ref{fig:KSC_classification_map}, ~\ref{fig:Hou_classification_map}, respectively. Specifically, we can see that the classification maps of the previous methods except SSRN have more or less noisy areas whether they belong to the conventional methods or the deep learning-based models. On the contrary, the classification maps of SSRN are smoother and the corresponding classification results of each class are more accurate apparently. However, our PDML structure generates a better quality of classification maps which not only get rid of irregular noise information but also precisely present the same color as the ground truth map for the testing samples, which further demonstrates the superior performance of our proposed framework.

\textbf{Visualization of embedding space:}
Furthermore, we visualize the embeddings before and after training with the t-SNE plots in Figure~\ref{fig:tsne} with different colors standing for different classes of embeddings. We observe that the embeddings are presented desultorily before the training procedure while the embeddings of the same class tend to gather together after training, which verifies that our PDML framework effectively reduces the intraclass distances and enlarges the interclass distances in the embedding space.

\section{Conclusion}
\setlength{\baselineskip}{12.33pt}
In this paper, we have presented a probabilistic deep metric learning (PDML) framework to improve the performance of hyperspectral image classification tasks.
We use the Gaussian distribution to model the label uncertainty of the spectral channel in the input patches and employ the Monte-Carlo sampling to generate various training samples from these distributions.
We then impose a probabilistic contrastive loss to enlarge interclass distances and decrease intraclass distances for better discriminative ability of the model.
We have performed diverse experiments on three widely-used hyperspectral datasets to examine the effectiveness of the proposed PDML framework, which achieved the state-of-the-art classification performance as well as enhanced the accuracy of various architectures.
It is interesting to employ the proposed framework in other directions such as object localization, change detection, and segmentation tasks in remote sensing as future works.

\ifCLASSOPTIONcaptionsoff
  \newpage
\fi

{\small

\bibliographystyle{ieee}
}

\vfill

\end{document}